\documentclass[final,2p,twocolumn]{elsarticle}
\usepackage{amssymb}
\usepackage[numbers]{natbib}
\usepackage{amsmath}
\usepackage{mathtools}
\usepackage{enumitem}
\usepackage{subfigure}
\usepackage[T1]{fontenc}
\usepackage[utf8]{inputenc}
\usepackage{lipsum} 

\usepackage{lineno}
\usepackage{hyperref}
\usepackage{multirow}
\usepackage{graphicx}
\usepackage{subcaption}
\usepackage{cleveref}
\usepackage[table,xcdraw]{xcolor}
\usepackage{booktabs}
\usepackage{float}

\makeatletter
\def\ps@pprintTitle{%
    \let\@oddfoot\@empty
    \let\@evenfoot\@empty
}
\makeatother

\begin{document}

\begin{frontmatter}

\title{Comprehensive framework for evaluation of deep neural networks in detection and quantification of lymphoma from PET/CT images: clinical insights, pitfalls, and observer agreement analyses}

\author[afl1,afl2,afl3]{Shadab Ahamed\corref{cor1}} 
\author[afl3]{Yixi Xu}
\author[afl2,afl4]{Sara Kurkowska}
\author[afl5]{Claire Gowdy}
\author[afl6]{Joo H. O}
\author[afl7]{Ingrid Bloise}
\author[afl7]{Don Wilson}
\author[afl7]{Patrick Martineau}
\author[afl7]{Fran\c{c}ois B\'{e}nard}
\author[afl2]{Fereshteh Yousefirizi}
\author[afl3]{Rahul Dodhia}
\author[afl3]{Juan M. Lavista}
\author[afl3]{William B. Weeks}
\author[afl1,afl2]{Carlos F. Uribe}
\author[afl1,afl2]{Arman Rahmim}
\cortext[cor1]{Corresponding author: Shadab Ahamed (email: shadab.ahamed@hotmail.com)}

\affiliation[afl1]{organization={University of British Columbia},
            city={Vancouver},
            state={BC},
            country={Canada}}
\affiliation[afl2]{organization={BC Cancer Research Institute},
            city={Vancouver},
            state={BC},
            country={Canada}}
\affiliation[afl3]{organization={Microsoft AI for Good Lab},
            city={Redmond},
            state={WA},
            country={USA}}
\affiliation[afl4]{organization={Pomeranian Medical University},
            city={Szczecin},
            state={Zachodniopomorskie},
            country={Poland}}
\affiliation[afl5]{organization={BC Children's Hospital},
            city={Vancouver},
            state={BC},
            country={Canada}}
\affiliation[afl6]{organization={St. Mary's Hospital},
            city={Seoul},
            country={Republic of Korea}}
\affiliation[afl7]{organization={BC Cancer},
            city={Vancouver},
            state={BC},
            country={Canada}}

\begin{abstract}
\textit{Purpose:} This study addresses critical gaps in automated lymphoma segmentation from PET/CT imaging, often overlooked in prior work. While deep learning has been applied to this task, few studies evaluate generalizability on external or out-of-distribution data. Similarly, intra- and inter-observer variability analyses remain rare, limiting understanding of task difficulty. Moreover, most methods emphasize global segmentation metrics, neglecting lesion-level characteristics that are crucial for clinical decision-making.\\
\textit{Methods:} We propose a clinically-relevant evaluation framework to assess four commonly used deep segmentation networks (ResUNet, SegResNet, DynUNet, SwinUNETR) on 611 PET/CT cases from multi-institutional datasets spanning varied lymphoma subtypes and lesion characteristics. In addition to the Dice similarity coefficient (DSC), we compute prediction errors on clinical lesion measures and analyze DSC performance as a function of these measures. Additionally, we use traditional lesion-specific detection criteria (1 and 2), providing insights into network's performance in identifying and localizing lesions respectively, and propose an additional Criterion 3 for segmenting lesions based on metabolic characteristics. Finally, we contextualize network performance by comparing it to expert human observers through intra- and inter-observer variability analyses.\\
\textit{Results:} Networks perform best on large, metabolically active lesions. Their error patterns closely resemble those of expert annotators, while small and faint lesions remain challenging for both networks and physicians.\\
\textit{Conclusion:} Our clinically-relevant benchmarking framework enables more consistent and meaningful evaluation of lymphoma segmentation models, supporting robust decision-making in patient care. The approach is extensible to other architectures and disease types. Code is available at: \url{https://github.com/microsoft/lymphoma-segmentation-dnn}.

\end{abstract}


\end{frontmatter}

\section{Introduction}
\label{sec:introduction}
Fluorodeoxyglucose ($^{18}$F-FDG) PET/CT imaging is the standard of care for lymphoma patients, providing diagnoses, staging, and therapy response evaluation. However, traditional qualitative assessments, like Deauville scores \cite{deauville_score}, can introduce variability due to observer's subjectivity in image interpretation. Using quantitative PET analysis that incorporates lesion measures such as mean lesion standardized uptake value ($\text{SUV}_\text{mean}$), total metabolic tumor volume (TMTV), and total lesion glycolysis (TLG) offers a promising path to more reliable prognostic decisions, enhancing our ability to predict patient outcomes in lymphoma with greater precision \cite{suvmean_tmtv_tlg}.

Quantitative assessment in PET/CT imaging often relies on manual lesion segmentation, which is time-consuming and prone to observer variabilities. Traditional thresholding-based automated techniques can miss low-uptake disease and produce false positives in regions of physiological high uptake of radiotracers. Therefore, deep learning offers promise for automating lesion segmentation, reducing variability, increasing patient throughput, and potentially aiding in the detection of challenging lesions \cite{breast_cancer}.

In particular, a class of deep networks called convolutional neural networks (CNNs) rely heavily on large, annotated datasets, as smaller datasets may hinder generalizability. Moreover, variability in lymphoma lesions in terms of size, shape, and metabolic activity complicates network training in the absence of well-defined priors. Models trained on inconsistent annotations due to observer variability can further perpetuate errors. Understanding these challenges is crucial towards harnessing the full potential of these methods in PET/CT quantitative analysis.

\section{Related work}
\label{sec:related_work}
Numerous works have explored the application of deep learning methods for segmenting lymphoma in PET/CT images. Yuan et al. \cite{feature_fusion} developed a feature fusion technique to utilize the complementary information from multi-modality data. Hu et al. \cite{fusion_2d_3d} proposed fusing a combination of 3D ResUNet trained on volumetric data and three 2D ResUNet trained on 2D slices from three orthogonal directions for enhancing segmentation performance. Li et al. \cite{densexnet} proposed DenseX-Net trained end-to-end and integrating supervised and unsupervised methods for lymphoma detection. Liu et al. \cite{neg_sample_aug_label_guidance} introduced techniques such as patch-based negative sample augmentation and label guidance for training a 3D Residual-UNet for lymphoma segmentation. A major limitation of all these works was that they were developed on relatively smaller-sized datasets (less than 100 images). Moreover, most of these methods did not compare the performance of their proposed methods with other baselines or with the performance of physicians.

Constantino et al. \cite{semiauto_deeplearning} compared the performances of 7 semi-automated and 2 deep learning segmentation methods, while Weisman et al. \cite{automated_eleven} compared 11 automated segmentation techniques, although these studies were performed on relatively small sized datasets of 65 and 90 cases, respectively. Weisman et al. \cite{weisman_deepmedic} compared the segmentation performances of automated 3D Deep Medic method with that of physician although even this study included just 90 lymphoma cases. Except for \cite{weisman_deepmedic}, none of these studies reported model generalization on out-of-distribution dataset (such as on data collected from different centers), limiting their robustness quantification and external validity. Jiang et al. \cite{Jiang2022} used a relatively larger dataset as compared to the above studies with 297 images to train a 3D UNet and performed out-of-distribution testing on 117 images collected from a different center. To the best of our knowledge, the largest lymphoma PET/CT dataset for deep learning-based lesion segmentation ever reported is the work by Blanc-Durand et al. \cite{largest_lymphoma_dataset} who used 639 images for model development and 94 for external testing; however, this study only used standard segmentation evaluation metrics and assessed their model's ability for predicting accurate TMTV. Both studies \cite{Jiang2022} and \cite{largest_lymphoma_dataset} were restricted to patients diagnosed with diffuse large B-cell lymphoma (DLBCL), thereby focusing solely on a specific lymphoma subtype.

Most existing studies on lymphoma segmentation report their performances on generic segmentation metrics such as Dice similarity coefficient (DSC), intersection-over-union (IoU), sensitivity, etc. In the presence of large segmented lesions, small missed lesions or small false positives do not significantly contribute to the DSC value. Hence, there is a need to report the volumes of false positives and false negatives. It will also be beneficial to evaluate the detection performances on a per-lesion basis (number of connected components detected vs missed), since detection of even a few voxels of all lesions can help physicians quickly locate the regions of interest, even if the DSC is low.

Busby et al. \cite{busby2018bias} highlighted how a spectrum of cognitive biases -- such as anchoring, availability, and satisfaction-of-search -- shape radiologists' perceptual and reasoning processes, making image interpretation inherently subjective. These biases systematically skew detection performance and drive substantial inter-reader variability, emphasizing that even expert annotations are colored by individual expectations and heuristics. Within the lymphoma expert segmentation literature, the difficulty of the segmentation/detection task is often not assessed via intra- or inter-observer agreement analysis. \\
\indent The segmentation performance of networks in PET often relies on various critical lesion measures, such as $\text{SUV}_\text{mean}$, $\text{SUV}_\text{max}$, and TMTV \cite{tumor_feature_performance_dependence}, which are usually poorer for small and faint lesions. However, existing studies usually do not report the distribution of lesion measures within privately-owned datasets, leading to a lack of transparency regarding dataset properties. This knowledge gap hinders understanding how model performance relates to data traits, thereby hindering scalability and clinical application. To address this issue, researchers must detail dataset information and assess how network performance depends on these traits.\\\\
\indent \textit{Contributions:} This study addresses critical gaps in the application of automated deep learning techniques for lymphoma segmentation from PET/CT images, emphasizing clinical robustness. Unlike prior work, our proposed framework incorporates external validation using multi-institutional datasets and performs comprehensive evaluations comparing the performance of deep segmentation networks (ResUNet, SegResNet, DynUNet, and SwinUNETR) to human experts. Key contributions of our work include: (i) rigorous assessment of model generalizability with external and out-of-distribution testing to enhance real-world clinical applicability; (ii) detailed analysis of segmentation performance beyond standard metrics (like DSC) to emphasize lesion-specific detection and clinically relevant metrics; (iii) detection criteria tailored for clinical lesion identification, incorporating both traditional measures and measures based on metabolic characteristics for evaluating networks lesion detection capabilities; (iv) contextualized performance comparisons with expert human annotators through intra- and inter-observer variability analyses, bridging the gap between automated methods and clinical practice; and (v) public availability of the implementation to enable reproducibility and further advancements in automated PET lesion segmentation research.


\section{Materials and methods}
\label{sec:methods}

\subsection{Dataset}
\label{subsec:dataset}
In this work, we used a large, diverse and multi-institutional whole-body PET/CT dataset with a total of 611 cases. These scans came from four retrospective cohorts: (i) \textit{DLBCL-BCCV}: 107 scans from 79 patients with DLBCL from BC Cancer, Vancouver (BCCV), Canada; (ii) \textit{PMBCL-BCCV}: 139 scans from 69 patients with PMBCL from BC Cancer; (iii) \textit{DLBCL-SMHS}: 220 scans from 219 patients with DLBCL from St. Mary's Hospital, Seoul (SMHS), South Korea; (iv) \textit{AutoPET lymphoma}: 145 scans from 144 patients with lymphoma from the University Hospital T\"{u}bingen, Germany \cite{Gatidis2022}. Additional description about the dataset from each cohort is given in \Cref{tab:cohort_data}. The cohorts (i)-(iii) are collectively referred to as the \textit{internal} cohort. The cohort (iv) was obtained from the AutoPET challenge dataset (public) \cite{Gatidis2022} and is referred to as the \textit{external} cohort. For all the cohorts, the PET and CT images were acquired simultaneously on a PET-CT scanner in a single session. As a result, the PET and CT images were anatomically aligned up to minor shifts due to physiological motion. The ethical statements about these datasets can be found in \Cref{ethical_statements}.

\begin{table*}[h]
\centering
\resizebox{0.95\linewidth}{!}{%
\Large
\begin{tabular}{ccccc}
\hline
\textbf{Cohort} &
  \textbf{Institution} &
  \textbf{\begin{tabular}[c]{@{}c@{}}Number of \\ patients/scans; \\ Age in years \\ (median \& range); \\ Sex\end{tabular}} &
  \textbf{Scanner Model} &
  \textbf{Reconstruction method} \\ \hline
\begin{tabular}[c]{@{}c@{}}DLBCL-BCCV\\ (internal)\end{tabular} &
  \begin{tabular}[c]{@{}c@{}}BC Cancer, \\ Vancouver, \\ Canada\end{tabular} &
  \begin{tabular}[c]{@{}c@{}}79/107;\\ 69 {[}19, 89{]};\\ M/F: 37/42\end{tabular} &
  \begin{tabular}[c]{@{}c@{}}CPS 1080: 5;\\ GE Medical Systems \\ (Discovery 600 / 690 / MI): \\ 43 / 58 / 1\end{tabular} &
  \begin{tabular}[c]{@{}c@{}}OSEM2D / VPFX / VPFXS / VPHDS: \\ 5 / 3 / 8 / 91\end{tabular} \\ \hline
\begin{tabular}[c]{@{}c@{}}PMBCL-BCCV\\ (internal)\end{tabular} &
  \begin{tabular}[c]{@{}c@{}}BC Cancer, \\ Vancouver, \\ Canada\end{tabular} &
  \begin{tabular}[c]{@{}c@{}}69/139\\ 33 {[}19, 79{]};\\ M/F: 30/39\end{tabular} &
  \begin{tabular}[c]{@{}c@{}}CPS 1080: 20; \\ GE Medical Systems \\ (Discovery 600 / 690): 55 / 64\end{tabular} &
  \begin{tabular}[c]{@{}c@{}}OSEM2D / VPFX / VPHD / VPHDS: \\ 20 / 1 / 6 / 112\end{tabular} \\ \hline
\begin{tabular}[c]{@{}c@{}}DLBCL-SMHS\\ (internal)\end{tabular} &
  \begin{tabular}[c]{@{}c@{}}St. Mary's Hospital, \\ Seoul, \\ South Korea\end{tabular} &
  \begin{tabular}[c]{@{}c@{}}219/220;\\ 61 {[}14, 87{]};\\ M/F: 117/102\end{tabular} &
  \begin{tabular}[c]{@{}c@{}}Siemens (1094 / \\ Biograph40-TruePoint): \\ 53 / 167\end{tabular} &
  \begin{tabular}[c]{@{}c@{}}OSEM2D / PSF: \\ 133 / 87\end{tabular} \\ \hline
\begin{tabular}[c]{@{}c@{}}AutoPET lymphoma \cite{gatidis2024results}\\ (external)\end{tabular} &
  \begin{tabular}[c]{@{}c@{}}University Hospital\\ T\"{u}bingen,\\ Germany\end{tabular} &
  \begin{tabular}[c]{@{}c@{}}144/145;\\ 47 {[}11, 85{]};\\ M/F: 76/88\end{tabular} &
  \begin{tabular}[c]{@{}c@{}}Siemens (Biograph128 / \\ Biograph128-mCT /  \\ SOMATOM Definition AS-mCT): \\ 3 / 129 / 13\end{tabular} &
  \begin{tabular}[c]{@{}c@{}}PSF+TOF: \\ 145\end{tabular} \\ \hline
\end{tabular}%
}
\caption{Patient and scan characteristics across four different lymphoma cohorts.}
\label{tab:cohort_data}
\end{table*}

\subsection{Ground truth (GT) annotations}
\label{subsec:ground_truth_annotation}
The DLBCL-BCCV, PMBCL-BCCV, and DLBCL-SMHS cohorts were separately segmented by three nuclear medicine physicians (referred to as Physicians 1, 3, and 4, respectively) from BC Cancer, Vancouver, BC Children's Hospital, Vancouver, and St. Mary's Hospital, Seoul, respectively. Additionally, another nuclear medicine resident (Physician 2) from BC Cancer segmented 35 cases from DLBCL-BCCV which were used for assessing inter-observer variability (\Cref{subsec:inter_observer_variability}). Physician 3 additionally re-segmented 60 cases from PMBCL-BCCV which were used for assessing intra-observer variability (\Cref{subsec:intra_observer_variability}). All expert annotations were performed using the semi-automatic gradient-based segmentation tool called PETEdge+ from the MIM workstation (MIM software, Ohio, USA). The AutoPET lymphoma data along with their GT segmentations \cite{Gatidis2022} were acquired from \textit{The Cancer Imaging Archive}. This dataset was annotated by two radiologists from the University Hospital T\"{u}bingen and the University Hospital of the LMU, Germany.

\subsection{Networks, tools and code}
\label{subsec:networks}
In this work, we trained four networks: (a) ResUNet \cite{residual_unet}, (b) SegResNet \cite{segresnet}, (c) DynUNet \cite{Isensee_2020}, and (d) SwinUNETR \cite{swinunetr}. The networks (a)-(c) are 3D CNNs, while (d) is a 3D transformer-based segmentation network. In particular, (b) was the winning solution in the HECKTOR 2022 challenge \cite{segresnet_hecktor}, and (c) is a 3D full-resolution implementation of nnUNet \cite{isensee2018nnunet}, the winner of the Medical Segmentation Decathlon 2018 challenge. Additionally, (a) and (c) were used by top performing teams in the AutoPET 2022-23 challenges \cite{gatidis2024results}, while (d) was one of the top performers in BraTS 2021 challenge \cite{swinunetr}. Hence, these networks are representative of the several state-of-the-art networks for lesion segmentation. Additional details on the architecture of these networks and the tools and training setups are briefly described in \Cref{app:network_architectures_and_training_setups}. The implementations have been open-sourced under the MIT License and can be found at: \url{https://github.com/microsoft/lymphoma-segmentation-dnn}.

\subsection{Training methodology} 
\label{subsec:training_methodology}
The data from internal cohorts (i) DLBCL-BCCV, (ii) PMBCL-BCCV, and (iii) DLBCL-SMHS (total 466 cases) were randomly split into training (302 cases), validation (76 cases), and internal test (88 cases) sets, while the (iv) AutoPET lymphoma cohort (145 cases) was used solely for external testing. The internal cohort split was stratified at patient level to avoid multiple images from the same patient being shared among training and validation/test sets. The models were first trained on the training set and the optimal hyperparameters were chosen on the validation set. Top models across different epochs were then tested on the internal and external test sets.  

Additional details for the training of these networks such as data preprocessing and augmentations, loss function, optimizer, learning rate scheduler, hyperparameters choices, sliding window inference method and postprocessing are described in \Cref{app:training_details}.

\subsection{Evaluation}
\label{subsec:evaluation_metrics}
\subsubsection{Segmentation metrics}
\label{subsubsec:segmentation_metrics}
To evaluate segmentation performance, we used patient-level foreground DSC, the volumes of false positive connected components that do not overlap with the GT foreground (FPV), and the volume of foreground connected components in the GT that do not overlap with the predicted mask (FNV) \cite{Gatidis2022}. For a foreground GT mask $G$ containing $L_\text{g}$ disconnected lesions $\{G_1, G_2, \ldots, G_{L_\text{g}}\}$ and the corresponding predicted foreground mask $P$ with $L_\text{p}$ lesions $\{P_1, P_2, \ldots, P_{L_\text{p}}\}$, these metrics are defined as, 
\begin{align}
    \text{DSC} &= 2\frac{|G \cap P|}{|G| + |P|}, \\
    \text{FPV} = v_p &\sum_{l=1}^{L_\text{p}} |P_l| \delta(|P_l \cap G|), \\
    \text{FNV} = v_g &\sum_{l=1}^{L_\text{g}} |G_l| \delta(|G_l \cap P|),
\end{align}
where $\delta(x):= 1$ for $x=0$ and $\delta(x):= 0$ otherwise; $v_g$ and $v_p$ represent the voxel volumes for GT and predicted mask, respectively (with $v_p = v_g$ for a given GT and predicted masks pair since the predicted mask was resampled to the original GT coordinates). 

\begin{figure}[h]
\centering
\includegraphics[width=\linewidth]{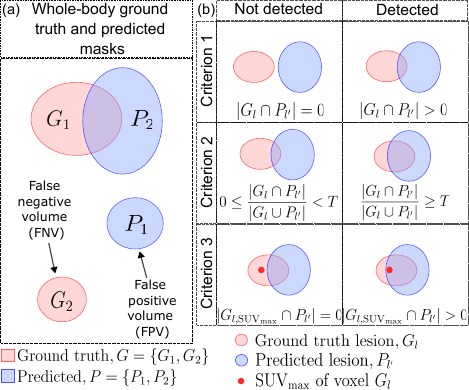}
\caption{(a) Illustration for the two segmentation metrics false positive volume (FPV) and false negative volume (FNV). (b) Illustration for defining a true positive detection via three criteria, as explained in \Cref{subsubsec:detection_metrics}.}
\label{fig:fpv_fnv_detection_metrics}
\end{figure}

We reported the median with interquartile range (IQR) for these metrics on the internal and external test sets. We also report mean DSC with standard deviation (SD). We chose to report median since the mean values were prone to outliers and our sample median was always higher (lower) for DSC (for FPV and FNV) than the sample mean. An illustration of FPV and FNV is given in \Cref{fig:fpv_fnv_detection_metrics} (a). 

\subsubsection{Detection metrics}
\label{subsubsec:detection_metrics}
Apart from the segmentation metrics, we also assessed model performance via three detection-based metrics for evaluating the detectibility of individual lesions within a patient. These metrics were utilized to address the clinical need for counting the lesions (Criterion 1), localizing them (Criterion 2), and segmenting lesions based on metabolic characteristic $\text{SUV}_\text{max}$ (Criterion 3). \\

In Criterion 1, a predicted lesion $P_{l^{\prime}}$ was labeled as true positive (TP) if at least one of
\begin{equation}
|G_l \cap P_{l^{\prime}}| > 0 \text{  for  } l \in \{1, 2,  \ldots, L_{g}\},
\end{equation}
otherwise $P_{l^{\prime}}$ was labeled as false positive (FP). Similarly, a GT lesion $G_l$ was labeled as false negative (FN) if
\begin{equation}
    |G_l \cap P_{l^{\prime}}| = 0 \text{  for all  } l^{\prime} \in \{1, 2, \ldots L_\text{p}\}.
\end{equation}
This definition is a weak detection criterion, where any predicted lesion is considered TP by having just one of its voxels overlap with any GT lesion. \\

In Criterion 2, all predicted lesions were first matched to their corresponding GT lesions by maximizing the IoU between each predicted and GT lesion pair. For a predicted lesion $P_{l^{\prime}}$ matched to a GT lesion $G_l$, $P_{l^{\prime}}$ was labeled as TP if 
\begin{equation}
\text{IoU}(G_l, P_{l^\prime}) = \frac{|G_l \cap P_{l^{\prime}}|}{|G_l \cup P_{l^{\prime}}|} \geq T,
\end{equation}
and as FP if 
\begin{equation}
    0 \leq \frac{|G_l \cap P_{l^{\prime}}|}{|G_l \cup P_{l^{\prime}}|} < T,
\end{equation}
where the threshold $T$ was set to 0.5. This was a stronger detection criterion than Criterion 1 as it required the matching of GT and predicted lesions and also imposed IoU $>$ 0.5 condition.\\

It has been shown in \cite{hirata2021preliminary} that lesion SUV$_\text{max}$ in FDG-PET images could be used as an identifier to localize the lesion. Based on that study, along with discussions with our own physicians, we introduce a third detection criterion. In Criterion 3, all predicted and GT lesions were first matched by maximizing IoU. For a matched pair $P_{l^{\prime}}$ and $G_l$, the lesion $P_{l^{\prime}}$ was labeled as TP if
\begin{equation}
|G_{l, \text{SUV}_\text{max}} \cap P_{l^{\prime}}| > 0,
\end{equation}
otherwise $P_{l^\prime}$ was labeled as FP, where $G_{l, \text{SUV}_\text{max}}$ is the voxel within $G_l$ with the maximum SUV from the corresponding PET image. Hence, this detection criterion results in higher sensitivity upon segmenting the $\text{SUV}_\text{max}$ voxel of the GT lesions.  For both Criteria 2 and 3, a GT lesion that was not matched with any of the predicted lesions based on IoU maximization was labeled as FN, while a predicted lesion not matched with any GT lesion was labeled as FP.
The clinical relevance of these lesion detection criteria have been summarized in \Cref{tab:detection_criteria_clinical_relevance}.

Although the definition for detection metrics FP and FN might appear similar to the segmentation metrics FPV and FNV, on careful investigation, they are not (\Cref{fig:fpv_fnv_detection_metrics} (a) and (b)). The metrics FPV and FNV are defined at the voxel level for each patient, while the detection metrics FP and FN in Criteria 1, 2, and 3 are defined on a per-lesion basis.

\begin{table*}[]
\centering
\resizebox{\textwidth}{!}{%
\begin{tabular}{@{}lll@{}}
\toprule
\textbf{Criteria} & \textbf{Description} & \textbf{Clinical relevance} \\ \midrule
Criterion 1 & \begin{tabular}[c]{@{}l@{}}A weak detection criterion that labels a lesion as\\  true positive (TP) if any voxel of the predicted \\ lesion overlaps with the ground truth lesion.\end{tabular} & \begin{tabular}[c]{@{}l@{}}Ensures that even the smallest part of a lesion is detected, \\ aiding physicians in quickly identifying potential regions \\ of interest, critical for early diagnosis.\end{tabular} \\ \midrule
Criterion 2 & \begin{tabular}[c]{@{}l@{}}A stronger detection criterion requiring an \\ intersection-over-union (IoU) \textgreater{}50\% between \\ predicted and ground truth lesions to label a TP.\end{tabular} & \begin{tabular}[c]{@{}l@{}}Ensures accurate lesion boundary delineation, critical for \\ precise treatment planning, especially in cases requiring \\ targeted interventions.\end{tabular} \\ \midrule
Criterion 3 & \begin{tabular}[c]{@{}l@{}}A clinical detection criterion labeling lesions \\ as TP if the predicted lesion contains the voxel \\ with the SUVmax.\end{tabular} & \begin{tabular}[c]{@{}l@{}}Emphasizes detection of metabolically active lesions, guiding \\ treatment by identifying potential targets and ensuring\\ key diagnostic markers are captured.\end{tabular} \\ \bottomrule
\end{tabular}%
}
\caption{Various detection criteria for defining a true positive detection among the predicted (segmented) lesions from networks (or physicians) used in this work. Criteria 1 and 2 are commonly used in computer vision. In this work, we propose Criterion 3 to emphasize and measure the detection performance based on metabolic characteristic of lesions in PET images.}
\label{tab:detection_criteria_clinical_relevance}
\end{table*}

\begin{table*}[]
\centering
\resizebox{\textwidth}{!}{%
\begin{tabular}{@{}lll@{}}
\toprule
\textbf{Lesion measure} &
  \textbf{Description} &
  \textbf{Clinical relevance} \\ \midrule
SUV$_\text{mean}$ &
  \begin{tabular}[c]{@{}l@{}}Average standardized uptake value (SUV) across \\ the segmented lesion(s).\end{tabular} &
  \begin{tabular}[c]{@{}l@{}}Reflects the metabolic activity of the lesions, which is \\ indicative of the tumor’s aggressiveness and response \\ to therapy. A decrease in SUV$_\text{mean}$ in FDG-PET \\ represents a decrease in the proliferative activity \\ of the tumor \cite{Ben-Haim88_suvmean}.\end{tabular} \\ \midrule
SUV$_\text{max}$ &
  Maximum SUV within the segmented lesion(s). &
  \begin{tabular}[c]{@{}l@{}}Serves as a marker for the most metabolically active\\ part of the tumor, useful for identifying hotspots and \\ guiding biopsies or further intervention \cite{hirata2021preliminary}.\end{tabular} \\ \midrule
Number of Lesions &
  Count of distinct lesions segmented in the scan. &
  \begin{tabular}[c]{@{}l@{}}Higher lesion count is associated with higher disease\\ burden and can guide treatment planning, including \\ systemic vs. localized therapies.\end{tabular} \\ \midrule
\begin{tabular}[c]{@{}l@{}}Total Metabolic \\ Tumor Volume \\ (TMTV)\end{tabular} &
  \begin{tabular}[c]{@{}l@{}}Summation of metabolic tumor volumes (MTVs) \\ across all lesions.\end{tabular} &
  \begin{tabular}[c]{@{}l@{}}An indicator of overall tumor burden; higher TMTV \\ correlates with poorer prognosis and can influence \\ risk stratification and treatment intensity \cite{tmtv, tmtv_tlg}.\end{tabular} \\ \midrule
\begin{tabular}[c]{@{}l@{}}Total Lesion \\ Glycolysis (TLG)\end{tabular} &
  \begin{tabular}[c]{@{}l@{}}Summation of the products of MTV and \\ SUVmean for all lesions, representing the\\ total glycolytic activity of the tumor burden.\end{tabular} &
  \begin{tabular}[c]{@{}l@{}}Captures both the volume and metabolic activity of \\ tumors, providing a composite metric for assessing \\ disease severity and treatment response \cite{tmtv_tlg, tmtv_tlg2}.\end{tabular} \\ \midrule
\begin{tabular}[c]{@{}l@{}}Lesion Dissemination, \\ D$_\text{max}$\end{tabular} &
  \begin{tabular}[c]{@{}l@{}}Maximum distance between two lesion \\ voxels in the segmented scan.\end{tabular} &
  \begin{tabular}[c]{@{}l@{}}Reflects the spatial spread of the disease, which can\\ impact staging, prognosis, and decisions about \\ systemic versus localized therapies \cite{dmax1, dmax2}.\end{tabular} \\ \bottomrule
\end{tabular}%
}
\caption{Various PET lesion measures and their clinical relevance. Additionally, these measures have been shown to be predictive of lymphoma patient outcomes in different studies \cite{suvmean_tmtv_tlg,review_paper}.}
\label{tab:lesion_measure_clinical_relevance}
\end{table*}

\begin{figure*}[h!]
\centering
\includegraphics[width=1\linewidth]{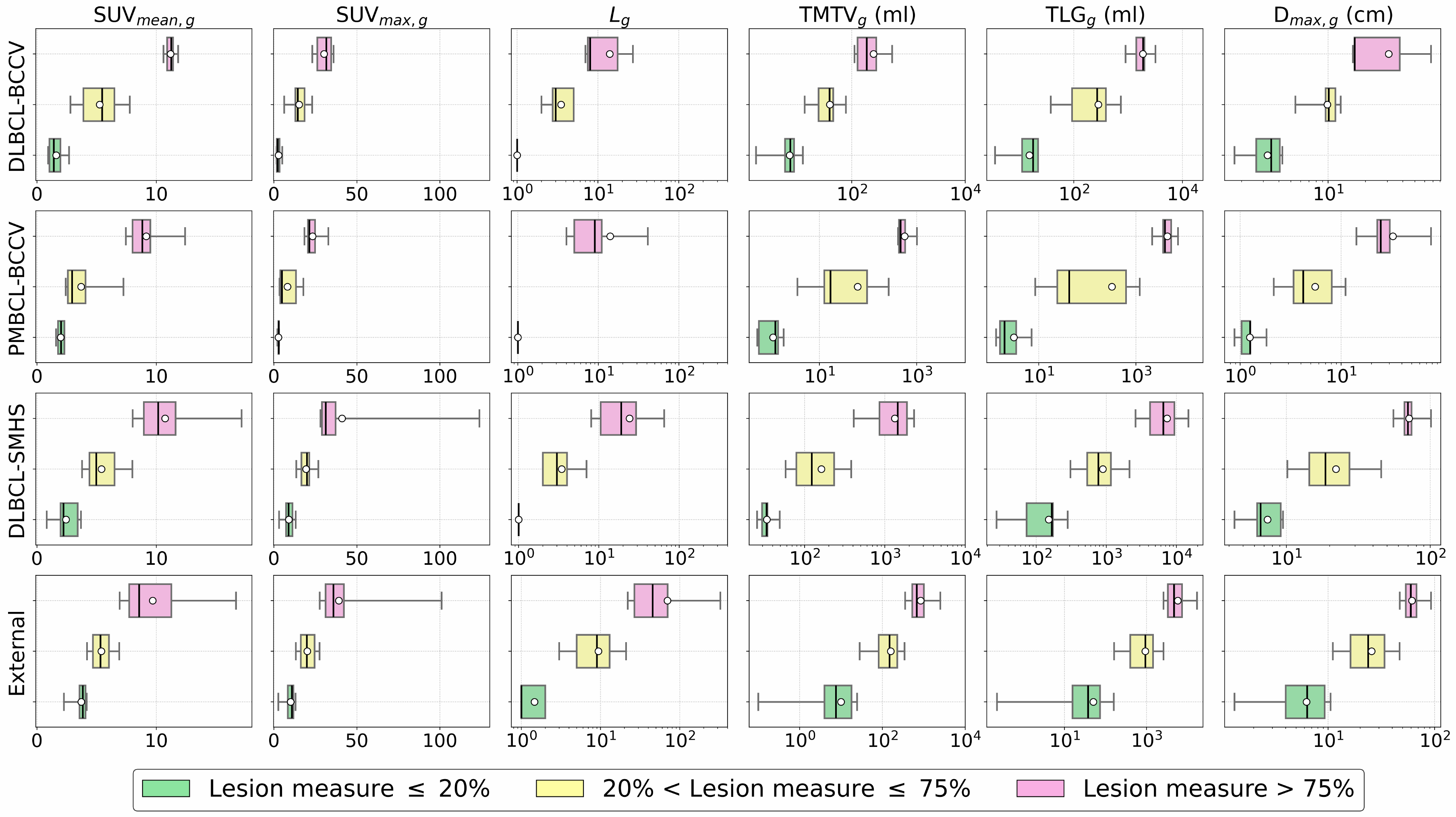}
\caption{Distribution of ground truth (GT) lesion measures on the test sets from different cohorts, showing the diversity of the datasets. The subscript $g$ denotes GT, i.e., these measures were extracted from the GT segmentation masks annotated by physicians. A cohort-level summary of these same measures -- aggregated into the internal (DLBCL-BCCV, PMBCL-BCCV, DLBCL-SMHS) and external (AutoPET) test sets -- has been shown in \Cref{tab:lesion_measures_internal_external} for quick numerical comparison.}
\label{fig:distribution_of_lesion_measures}
\end{figure*}

\subsubsection{Clinically-relevant lesion measures and intra- and inter-observer agreement analysis}
\label{subsubsec:clinically_relevant_biomarkers_intra_and_inter_observer_agreement_analysis}
For evaluating lesion segmentation algorithms, incorporating clinically-relevant lesion measures provides a comprehensive assessment that extends beyond traditional overlap-based metrics such as DSC. While standard metrics offer insights into the quality of segmentation itself, they may not always highlight the significance of the segmentation measures \cite{relaince_paper, review_paper}. Hence, in addition to evaluating traditional segmentation accuracy, we evaluated six clinically-relevant lesion measures: (i) patient-level lesion $\text{SUV}_\text{mean}$ and (ii) $\text{SUV}_\text{max}$, (iii) number of lesions, $L$, (iv) TMTV, (v) TLG, and (vi) lesion dissemination, $\text{D}_\text{max}$ from the GT and predicted masks. We defined $\text{D}_\text{max}$ as the distance between the two farthest located foreground voxels in cm. These measures are well-established prognostic biomarkers in lymphoma and have been shown to correlate strongly with patient outcomes, influencing treatment planning and risk stratification \cite{suvmean_tmtv_tlg, review_paper}. As emphasized in the RELAINCE framework \cite{relaince_paper}, such quantitative clinical readouts fall under task-based evaluation, which moves beyond proof-of-concept validations and instead focuses on the model’s ability to replicate actionable clinical outputs. By incorporating this evaluation strategy, we aim to assess not only how well networks segment lesions, but also how reliably they reproduce lesion-derived metrics that are critical for real-world clinical decision-making. To enhance the interpretability and clinical utility of these metrics, we also provide a comprehensive mapping that links each lesion measure to its clinical implications, as detailed in \Cref{tab:lesion_measure_clinical_relevance}.


Assessing the reproducibility of  measures enhances the confidence in the clinical utility of algorithm. We conducted equivalence testing via two one-sided test (TOST) \cite{lakens2018equivalence} to determine whether the means of predicted and GT lesion measures are close enough to be considered equivalent (\Cref{subsubsec:reproducibility_of_lesion_measures}). For a lesion measure $b$, the \% mean difference between the predicted ($b_p$) and GT ($b_g$) lesion measures was computed as $100 \times \mathbb{E}[(b_p - b_g)/b_g] \%$. At a margin of $\Delta$, the lower $p$-value $p_l$ was computed to reject the possibility that the true \% mean relative difference is lower than $-\Delta$. Similarly, an upper $p$-value $p_h$ was computed to reject the possibility that the true \% mean difference is greater than $\Delta$. 

To obtain an evidence for equivalence, we must reject the null hypothesis (representing non-equivalence) if both $p_l < \alpha$ and $p_h < \alpha$. The test was conducted at a significance level $\alpha = 0.05$ and $\Delta = 20\%$. The region $[-\Delta, \Delta]$ is referred to as the region of \textit{clinical equivalence} and the 95\% confidence interval (CI) for the \% mean difference lying entirely within this region provides evidence for equivalence (showing reproducibility). The equivalence margin of $\Delta = 20\%$ was determined via discussion with physicians to reflect a clinically meaningful yet conservative tolerance for quantitative agreement in FDG PET/CT. Multicenter test–retest studies have reported repeatability errors for SUV-based metrics on the order of approximately 20–30\% even under standardized acquisition and analysis conditions, reflecting combined biological and technical variability \cite{hofheinz2017test}. Established response frameworks such as PERCIST \cite{wahl2009recist} similarly use thresholds of approximately $\pm 30\%$ change in SUV to define clinically relevant metabolic response. In this context, the selected $\pm 20\%$ margin represents a stringent criterion for quantitative equivalence, ensuring that agreement - if demonstrated - would exceed expected clinical measurement variability.
Furthermore, to quantify errors (predicted - GT) in the estimation of lesion measures as a function of GT measure values, we also plotted modified Bland-Altman plots for each network on the test set, as presented in \Cref{subsubsec:reproducibility_of_lesion_measures}.

We also carried out equivalence testing on lesion measures to quantify intra-observer variability involving two annotations made by Physician 3 on the same set of 60 cases (\Cref{subsec:intra_observer_variability}). To quantify inter-observer variability (\Cref{subsec:inter_observer_variability}), we computed mean DSC between the annotations by Physicians 1 and 2 on the same set of 35 cases and also mean DSC of each physician against the generated STAPLE \cite{staple} consensus between them. To quantify the agreement in lesion measures from the two annotations, we computed the inter-class correlation coefficient (ICC) showing the agreement between the measure values extracted from the annotations by two physicians.

\subsubsection{Lesion measure threshold analysis}
\label{subsubsec:lesion_measure_threshold_analysis}
We evaluated the DSC performance of the four networks on subsets of the combined internal and external test sets (\Cref{subsubsec:dsc_vs_biomarkers_combined}). Let the set $\mathcal{B}$ of patient level GT lesion measure $b$ on the test set be represented by $\mathcal{B} \coloneqq \{b_i\}_{i=1}^{N_\text{cases}}$, where $N_\text{cases} = 233$ is the total number of cases in the combined test set and $i$ denotes the case index. For a specific lesion measure $b$, subsets $\mathcal{B}_\text{sub}(t_b)$ of the test cases were created by selecting a threshold $t_b$ such that
\begin{equation}
\mathcal{B}_\text{sub}(t_b) \coloneqq \{b_i \mid b_i \geq t_b\}.
\end{equation}
For each value of $t_b$, median DSC was only evaluated on the subset $\{i | b_i \geq t_b\}$ of total test cases. The values of $t_b$ were chosen in the range $\mathcal{B}_{0} \leq t_b \leq \mathcal{B}_{85}$ with a step-size of $\Delta t_b$, where $\mathcal{B}_{0} = \text{min}(\mathcal{B})$ and $\mathcal{B}_{85}$ respectively represent the 0$^\text{th}$ quantile (i.e.,~the minimum) and the 85$^\text{th}$ quantile of the set $\mathcal{B}$. The value of $\Delta t_b$ (chosen appropriately based on the range of values for a specific GT lesion measure) for $\text{SUV}_\text{mean, g}$, $\text{SUV}_\text{max, g}$, $L_\text{g}$, $\text{TMTV}_\text{g}$, $\text{TLG}_\text{g}$, and $\text{D}_\text{max, g}$ were chosen to be 1, 2, 1, 25 ml, 150 ml, and 3 cm, respectively, where the subscript $\text{g}$ is used to indicate the GT lesion measures.

\section{Results}
\label{sec:results}
\subsection{Segmentation performance}
\label{subsec:segmentation_performance}

The performance of the four networks were first evaluated using the segmentation metrics as detailed in \Cref{subsubsec:segmentation_metrics}. In cases of ties, preference was given to the network with the lower IQR, indicative of lower variance. 

The SegResNet had the highest median DSC on both internal and external test sets with medians of 0.76 [0.27, 0.88] and 0.68 [0.40, 0.78], respectively. For the individual cohorts within the internal test set, ResUNet had the best DSC on both DLBCL-BCCV and PMBCL-BCCV with a median of 0.72 [0.24, 0.89] and 0.74 [0.02, 0.90], respectively, while SegResNet had the best DSC of 0.78 [0.62, 0.87] on DLBCL-SMHS. SegResNet also had the best FPV on both internal and external test sets with values of 4.55 [1.35, 31.51] ml and 21.46 [6.30, 66.44] ml. Despite ResUNet winning on DSC for DLBCL-BCCV and PMBCL-BCCV sets, SegResNet had the best FPV on both these sets with median values of 5.78 [0.61, 19.97] ml and 2.15 [0.52, 7.18] ml, respectively, while ResUNet had the best FPV of 8.71 [1.19, 34.1] ml on DLBCL-SMHS. Finally, SwinUNETR had the best median FNV of 0.0 [0.0, 4.65] ml on the internal test set, while ResUNet had the best median FNV of 0.41 [0.0, 3.88] ml on the external test set. On DLBCL-BCCV and DLBCL-SMHS, SwinUNETR had the best median FNV of 0.09 [0.0, 3.39] ml and 0.0 [0.0, 8.83] ml, respectively, while on PMBCL-BCCV, ResUNet, DynUNet, and SwinUNETR were tied, each with a median value of 0.0 [0.0, 1.24] ml.

A detailed quantitative segmentation performances of the four networks on the various test sets have been discussed in \Cref{subsubsec:quantitiative_comparison_between_networks} and \Cref{tab:final_segmentation_results}. Some qualitative segmentation visualizations are shown in \Cref{fig:images_similar_performance,fig:images_dissimilar_performance}.

\subsubsection{Reproducibility of lesion measures}
\label{subsubsec:reproducibility_of_lesion_measures}

The lesion measures were computed at the patient level from the GT and predicted masks from each of the networks on the internal and external test sets. Plots showing the distribution of lesion measures on each of the test cohorts have been presented in \Cref{fig:distribution_of_lesion_measures}. The results of the equivalence tests have been presented in \Cref{fig:equivalence_testing} (i) and (ii) for the internal and external test sets respectively in \Cref{subsubsec:assessing_the_reproducibility_of_lesion_measures_via_equivalence_testing}.

\begin{figure}[h]
\centering
\includegraphics[width=\linewidth]{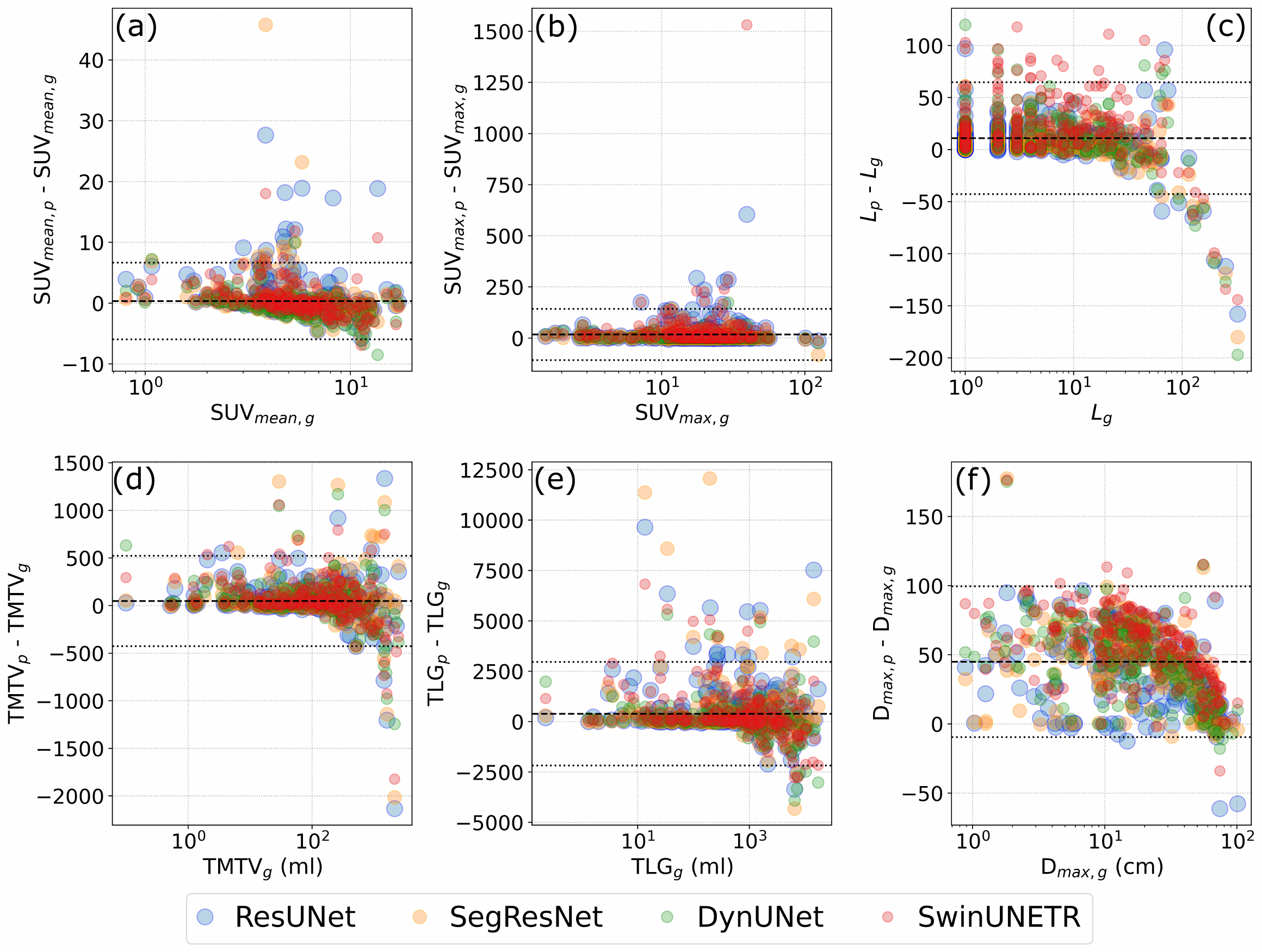}
\caption{Modified Bland-Altman plots showing errors (predicted - GT) in the estimation of lesion measures as a function of GT lesion measure values for the four networks, ResUNet, SegResNet, DynUNet, and SwinUNETR on the combined internal and external test set ($N_\text{cases} = 233$). The black dashed line represents the mean error over all networks and the black dotted lines represent $\pm 1.96$ SD on mean. These bounds are used for descriptive visualization of agreement and heteroscedasticity; formal equivalence was assessed separately using the TOST procedure with a predefined $\pm20\%$ equivalence margin, as shown in \Cref{fig:equivalence_testing}}. The $x$-axis has been represented on log scale.
\label{fig:modified_bland_altman_plot}
\end{figure}

The Modified Bland-Altman plots (\Cref{fig:modified_bland_altman_plot}) for each of the lesion measures show that although a large majority of errors lie within the $\pm 1.96$ SD of the mean error for all networks, there are several cases where the networks highly underestimate or overestimate lesion measure values by large amounts. For instance, $\text{SUV}_\text{max}$ estimation was largely within the $\pm 1.96$ SD of mean, except for a very few cases for which $\text{SUV}_\text{max}$ values were overestimated. Similarly, the errors on $\text{SUV}_\text{mean}$ were within $\pm 1.96$ SD of mean as well, except for a few cases where the values were highly overestimated. There were quite a few cases where the number of lesions was overestimated for $L_\text{g} < 10$, while they were underestimated for cases with $L_\text{g} > 100$. Similarly, for TMTV and TLG, there were a lot of of outliers with over- and underestimation for $\text{TMTV}_\text{g} > 100$ ml and $\text{TLG}_\text{g} > 10$ ml, respectively. For $\text{D}_\text{max}$, most of the error values were within the $\pm 1.96$ SD range of mean, although the mean of errors over all networks itself was very high ($\approx 50$ cm), showing an overestimation of $\text{D}_\text{max}$, which is consistent with the analyses in \Cref{fig:equivalence_testing} (i) and (ii).

\subsubsection{Effect of GT lesion measures values on network performance}
\label{subsubsec:dsc_vs_biomarkers_combined}
\begin{figure*}[h]
\centering
\includegraphics[width=0.9\linewidth]{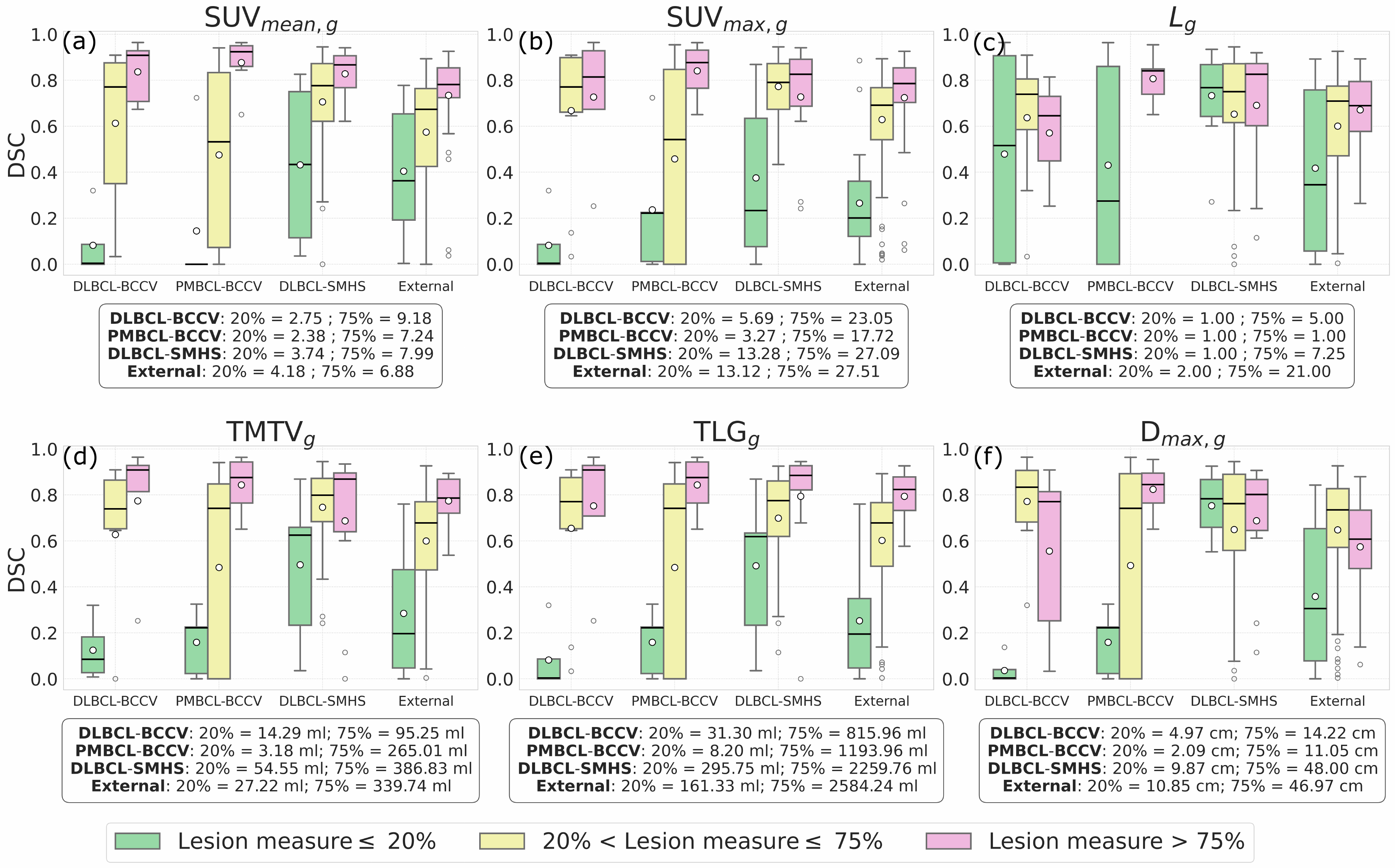}
\caption{\textbf{SegResNet} performance (DSC) distribution for different GT lesion measures on various test sets. For each test set, the DSC measure distributions have been presented as boxplots in three different categories, (i) Lesion measure $\leq$ 20\%tile, (ii) 20\%tile $<$ Lesion measure $\leq$ 75\%tile, (iii) Lesion measure $>$ 75\%tile. The mean and median values for each box have been represented as white circles and black horizontal lines, respectively. The boxes below each plot show the value of the 20\%tile and 75\%tile lesion measure on each of the test sets. Additional plots for ResUNet, DynUNet and SwinUNETR have been presented in \Cref{fig:lesion_measures_segregated_dsc_metrics_unet,fig:lesion_measures_segregated_dsc_metrics_dynunet,fig:lesion_measures_segregated_dsc_metrics_swinunetr} respectively in \Cref{subsubsec:lesion_measure_segregated_dsc_for_other_networks}.}
\label{fig:lesion_measures_segregated_dsc_metrics_segresnet}
\end{figure*}

First, we computed lesion measures for the internal and external test sets based on the GT lesion masks. For each lesion measure, the images in the internal and external cohorts were grouped based the values of lesion measures into three categories: (i) Lesion measure $\leq$ 20\%tile, (ii)  20\%tile $<$ Lesion measure $\leq$ 75\%tile, (iii) Lesion measure $>$ 75\%tile. For a particular network, the DSC distribution in each of these categories was accessed and plotted. \Cref{fig:lesion_measures_segregated_dsc_metrics_segresnet} shows the distribution of DSC for SegResNet segregated by the test cohorts within each lesion measure category for all lesion measures. From \Cref{fig:lesion_measures_segregated_dsc_metrics_segresnet} (a)-(b), it is evident that the categories with higher (mean and median) DSCs also had higher (mean and median) $\text{SUV}_\text{mean}$ and $\text{SUV}_\text{max}$ for both internal and external test sets. A similar trend was observed for TMTV and TLG (\Cref{fig:lesion_measures_segregated_dsc_metrics_segresnet} (d)-(e)), where categories with higher TMTV or TLG had higher DSCs as compared to categories with lower TMTV or TLG.  A similar trend was observed for $L$ (\Cref{fig:lesion_measures_segregated_dsc_metrics_segresnet} (c)) only on the external test set, but not on any of the internal test cohorts. Note that the 75\%tile $L$ on the external test set was considerably higher than any of the internal test sets. Finally, for $\text{D}_\text{max}$ (\Cref{fig:lesion_measures_segregated_dsc_metrics_segresnet} (f)), the category $\text{D}_\text{max} \leq 20\%\text{tile}$ had the lowest distribution of DSC (except on the DLBCL-SMHS set) that represents the set with lower spread of the disease, which can either correspond to cases with just one small lesion (harder to segment), or several lesions located in close proximity. Similar trends were observed for all other networks, ResUNet, DynUNet and SwinUNETR and their plots have been presented in \Cref{fig:lesion_measures_segregated_dsc_metrics_unet,fig:lesion_measures_segregated_dsc_metrics_dynunet,fig:lesion_measures_segregated_dsc_metrics_swinunetr} respectively in \Cref{subsubsec:lesion_measure_segregated_dsc_for_other_networks}.

\begin{figure}[h]
\centering
\includegraphics[width=1\linewidth]{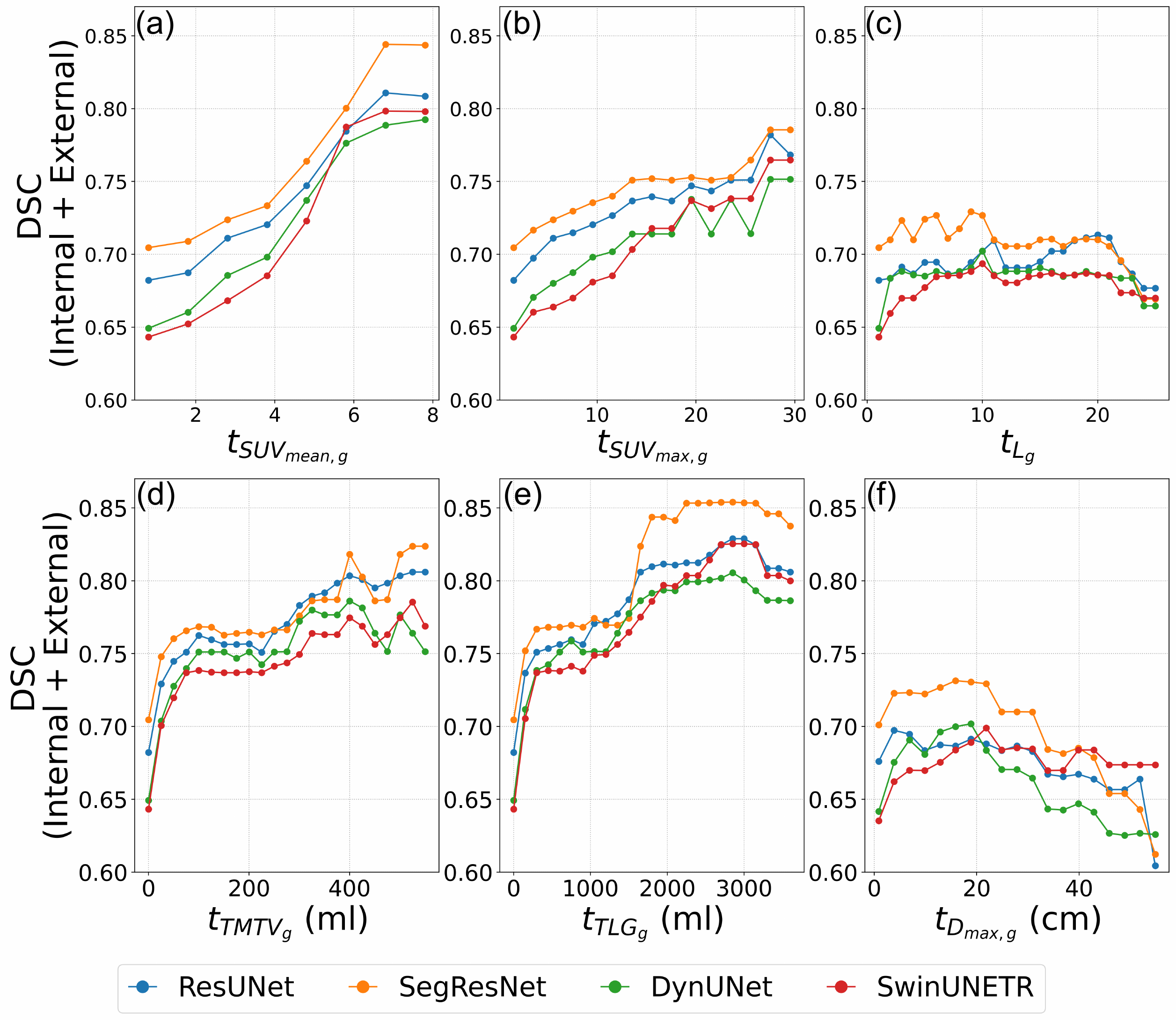}
\caption{The effect of GT lesion measure values from subsets of test cases (internal and external combined with $N_\text{cases} = 233$) on network performance. For a lesion measure $b$, a threshold $t_b$ was chosen and a subset of internal test cases were selected with $b \geq t_b$ and median DSC was computed on this subset. The value of $t_b$ were chosen in the range $[\mathcal{B}_0, \mathcal{B}_{85}]$ at steps of $\Delta t_b$, where $\mathcal{B}_0$ and $\mathcal{B}_{85}$ represent the 0$^\text{th}$ and 85$^\text{th}$ quantile of the set of all lesion measures on the internal test set, $\mathcal{B} = \{b_i\}_{i=1}^{N_\text{cases}}$. (a), (b), (d), and (e) show that, in general, the performance of networks increase on subset with larger values of $\text{SUV}_\text{mean}$, $\text{SUV}_\text{max}$, TMTV, and TLG, respectively up to certain values of $t_b$ (after which the performance plateaus), while for the number of lesions (c) and $\text{D}_\text{max}$ (e) this increase isn't very prominent.}
\label{fig:dsc_vs_gt_biomarkers_combined}
\end{figure}

We also performed lesion measure threshold analyses as explained in \Cref{subsubsec:lesion_measure_threshold_analysis} for each lesion measure and the results have been presented in \Cref{fig:dsc_vs_gt_biomarkers_combined}. For the patient level GT $\text{SUV}_\text{mean}$, the median DSC increases monotonically as a function of $t_{\text{SUV}_\text{mean, g}}$ for all networks up to $t_{\text{SUV}_\text{mean, g}} \approx 6$, while for $t_{\text{SUV}_\text{mean, g}} > 6$, the median DSC nearly plateaus (\Cref{fig:dsc_vs_gt_biomarkers_combined} (a)). The increase in median DSC up to $t_{\text{SUV}_\text{mean, g}} \approx 6$ is quite prominent, with an overall increase of about 13\%, 14\%, 14\%, and 16\% for ResUNet, SegResNet, DynUNet, and SwinUNETR, respectively.  Also, as a function of  $t_{\text{SUV}_\text{mean, g}}$, SegResNet had the best performance followed by ResUNet and other networks. DynUNet had better performance than SwinUNETR up to $t_{\text{SUV}_\text{mean, g}} \approx 5.5$ after which the trend between these two networks was flipped. From these analyses, it can be concluded that the networks are more likely to segment lesions accurately and return a higher median DSC on a set of cases with higher $\text{SUV}_\text{mean, g}$ values.

A similar trend was observed for $\text{SUV}_\text{max, g}$, where for all networks, the median DSC increased up to $t_{\text{SUV}_\text{max, g}} \approx 30$ with SegResNet performing the best, followed by ResUNet. Similar to the $\text{SUV}_\text{mean, g}$ measure, DynUNet performed better than SwinUNETR up to $t_{\text{SUV}_\text{max, g}} \approx 15$ after which their performances were flipped. The networks had an overall increase of about 10\%, 8\%, 10\%, and 12\%, respectively (\Cref{fig:dsc_vs_gt_biomarkers_combined} (b)). This shows that a higher $\text{SUV}_\text{max, g}$ also, in general, leads to better segmentation performance. For the number of lesions, there was no significant increase (all networks $<5$\%) as a function of $t_{L_\text{g}}$, while for  $\text{D}_\text{max, g}$, there was a slight initial increase (up to $t_{\text{D}_\text{max, g}} \approx 20 \text{ cm}$) but overall decrease later of 9\%, 12\%, 8\%, and 6\% for the networks, respectively. This suggests that while the number of lesions has no effect on performance, an increase in $\text{D}_\text{max, g}$ actually hurts the segmentation performance (\Cref{fig:dsc_vs_gt_biomarkers_combined} (c) and (f)). The median DSC also had considerable increase of 12\%, 12\%, 14\%, and 14\% for the networks respectively as a function of $t_{\text{TMTV}_\text{g}}$ with SegResNet and ResUNet with overlapping performances followed by DynUNet and SwinUNETR (\Cref{fig:dsc_vs_gt_biomarkers_combined} (d)), signifying that larger lesions are in general easier to segment accurately. Finally, a similar increase (up to $t_{\text{TLG}_\text{g}} \approx 2000 \text{ ml}$) and plateauing behavior in median DSC was observed for the measure TLG with an overall increase of about 15\%, 15\%, 16\%, and 18\% respectively for these networks, meaning that metabolically active large lesions are much easier to segment as compared to metabolically faint or smaller lesions (\Cref{fig:dsc_vs_gt_biomarkers_combined} (e)).

\subsection{Detection performance}
\label{subsec:detection_performance}
\begin{figure}[h]
\centering
\includegraphics[width=1\linewidth]{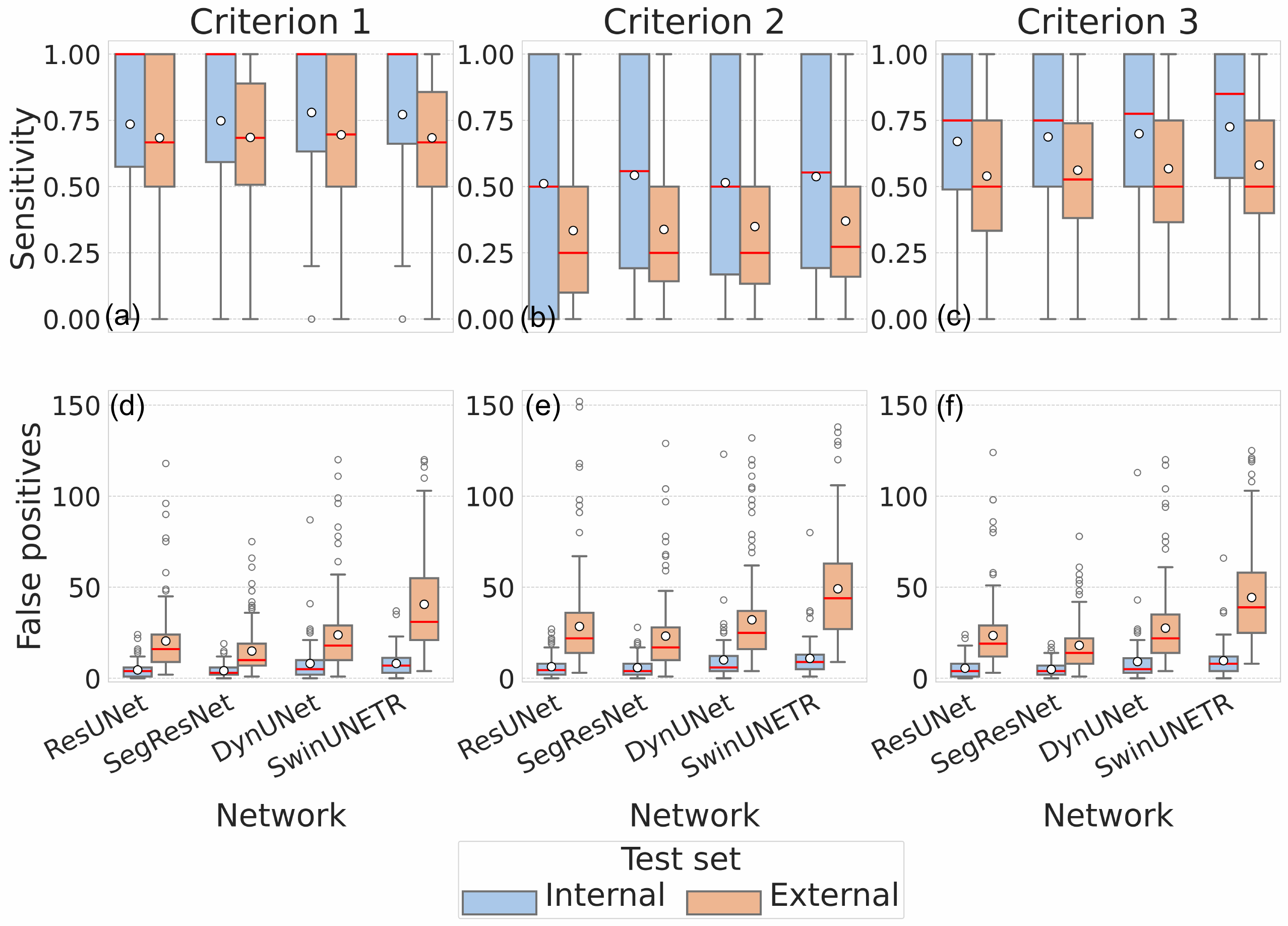}
\caption{Median detection sensitivity and FP per patient via the three detection criteria for the four networks on the internal and external test sets. The top and bottom edges of the boxes span the IQR, while the red horizontal lines and white circles represent the median and mean respectively. The whiskers length is set to 1.5 times IQR and the outliers have been shown as black diamonds.}
\label{fig:detection_boxplots}
\end{figure}

We evaluated network performance using the three detection criteria, as defined in \Cref{subsubsec:detection_metrics}. Criterion 1, being the weakest criterion, had the best overall detection sensitivity of all criteria across all networks on both internal and external test sets, followed by Criterion 3 and then Criterion 2 (\Cref{fig:detection_boxplots}). From Criterion 1, ResUNet, SegResNet, DynUNet, and SwinUNETR obtained median sensitivities of 1.0 [0.57, 1.0], 1.0 [0.59, 1.0], 1.0 [0.63, 1.0], and 1.0 [0.66, 1.0] respectively on internal test set, while on external set, they obtained 0.67 [0.5, 1.0], 0.68 [0.51, 0.89],  0.70 [0.5, 1.0], and 0.67 [0.5, 0.86] respectively. Naturally, there was a drop in performance upon going from internal to external testing. Furthermore, Criterion 1 gave the best network performance on the number of FP metrics with the networks obtaining 4.0 [1.0, 6.0], 3.0 [2.0, 6.0], 5.0 [2.0, 10.0], and 7.0 [3.0, 11.25] median FPs respectively on internal test set, and 16.0 [9.0, 24.0], 10.0 [7.0, 19.0], 18.0 [10.0, 29.0], and 31.0 [21.0, 55.0] median FPs respectively on external test set. \\

Furthermore, being a harder detection criterion, Criterion 2 had the lowest detection sensitivities for all networks with median being 0.5 [0.0, 1.0], 0.56 [0.19, 1.0], 0.5 [0.17, 1.0], and 0.55 [0.19, 1.0] respectively on internal test set, and 0.25 [0.1, 0.5], 0.25 [0.14, 0.5], 0.25 [0.13, 0.5], and 0.27 [0.16, 0.5] respectively on external test set. For this criterion, the drop in median sensitivities on going from the internal to external testing was comparable to those of Criterion 1. Similarly, for this criterion, the median FPs per patient were 4.5 [2.0, 8.0], 4.0 [2.0, 8.0], 6.0 [4.0, 12.25], and 9.0 [5.0, 13.0] respectively on internal test set, and 22.0 [14.0, 36.0], 17.0 [10.0, 28.0], 25.0 [16.0, 37.0], and 44.0 [27.0, 63.0] respectively on external test set. Despite the sensitivities being lower than in Criterion 1, the FPs per patient were similar on both internal and external test sets for Criterion 2.

Finally, the Criterion 3, based on the detection of the $\text{SUV}_\text{max}$ voxel of the lesions, was an intermediate criterion between Criteria 1 and 2, since the model's ability to detect lesions accurately increases with the lesion $\text{SUV}_\text{max}$ (\Cref{subsubsec:dsc_vs_biomarkers_combined}). For this criteria, the networks had median sensitivities of 0.75 [0.49, 1.0], 0.75 [0.5, 1.0], 0.78 [0.5, 1.0], and 0.85 [0.53, 1.0] respectively on the internal test set, and 0.5 [0.33, 0.75], 0.53 [0.38, 0.74], 0.5 [0.37, 0.75], and 0.5 [0.4, 0.75] respectively on the external test set. The drop in sensitivities between internal and external test sets is comparable to the other two criteria. Similarly, the networks had median FP per patient of 4.0 [1.0, 8.0], 4.0 [2.0, 7.0], 5.0 [3.0, 11.0], and 8.0 [4.0, 12.0] respectively on the internal test set, and 19.0 [12.0, 29.0], 14.0 [8.0, 22.0], 22.0 [14.0, 35.0], and 39.0 [25.0, 58.0] respectively on the external test set.

\subsection{Physician performances and comparisons to networks}
\label{subsec:intra_observer_variability}
\subsubsection{Intra-observer variability and reproducibility of lesion measures}
\label{subsubsec:intra_observer_variability_and_reproducibility_of_lesion_measures}

\begin{figure}[h]
\includegraphics[width=1\linewidth]{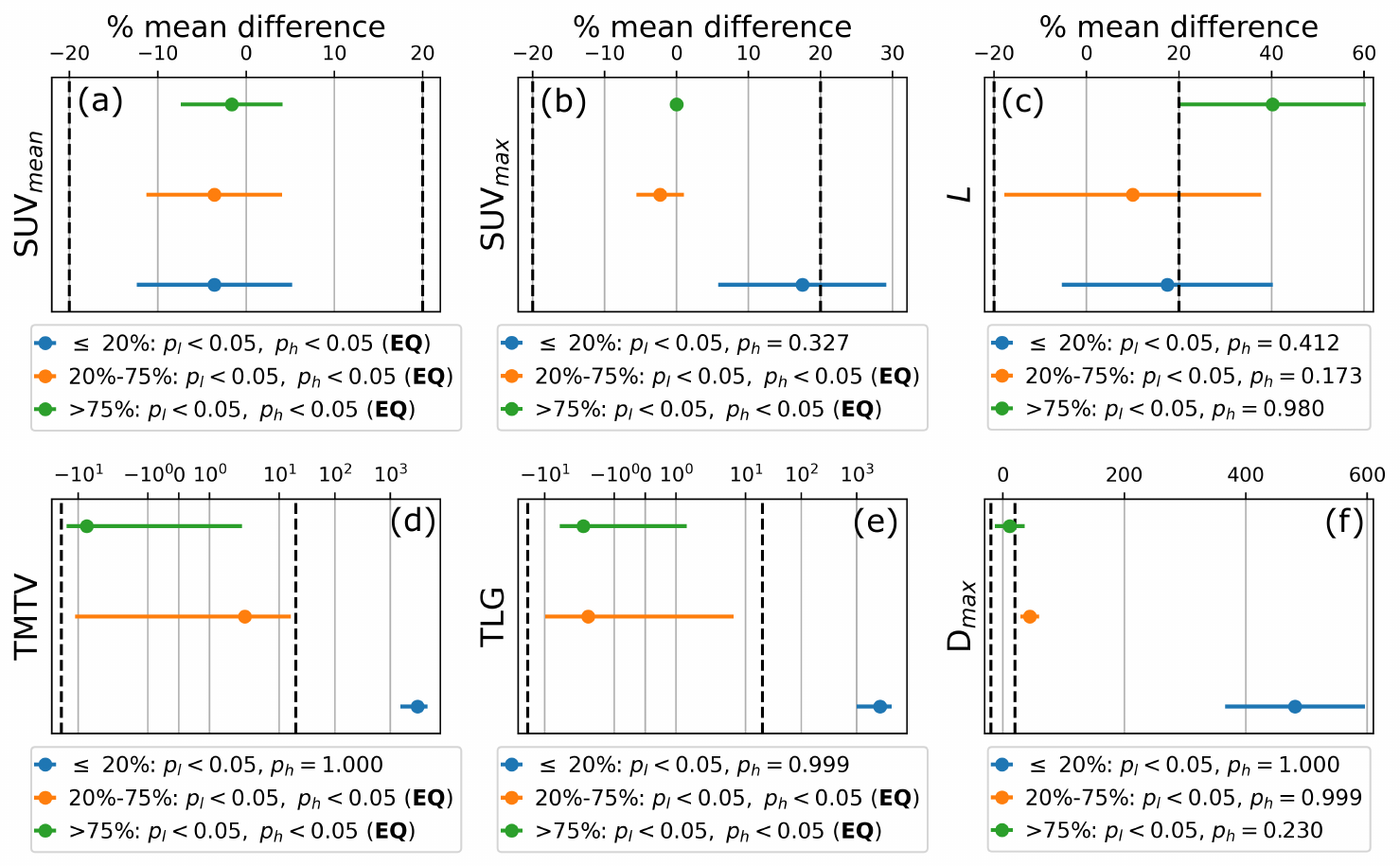}
\caption{Quantifying intra-observer (Physician 3) agreement for the reproducibility of lesion measures using equivalence testing by evaluating the \% mean difference between the original and re-segmented (new) GT lesion measures at $\alpha = 0.05$. The analyses were performed on 60 cases from the PMBCL-BCCV cohort which were segregated into three categories based on the value of lesion measures $\leq 20\%$tile,  20\%-75\%tile and $>$75\%tile (these thresholds were chosen on the PMBCL-BCCV test set for all lesion measures except for $L$ where it was chosen on this set of 60 cases). The horizontal axis for the TMTV and TLG plots have been shown on symlog scale. The label \textbf{EQ} represents equivalence, signifying reproducibility.}
\label{fig:equivalence_testing_intraobs_analysis_segregated_in_three_categories}
\end{figure}

\begin{figure*}[h!]
\centering
\includegraphics[width=0.9\textwidth]{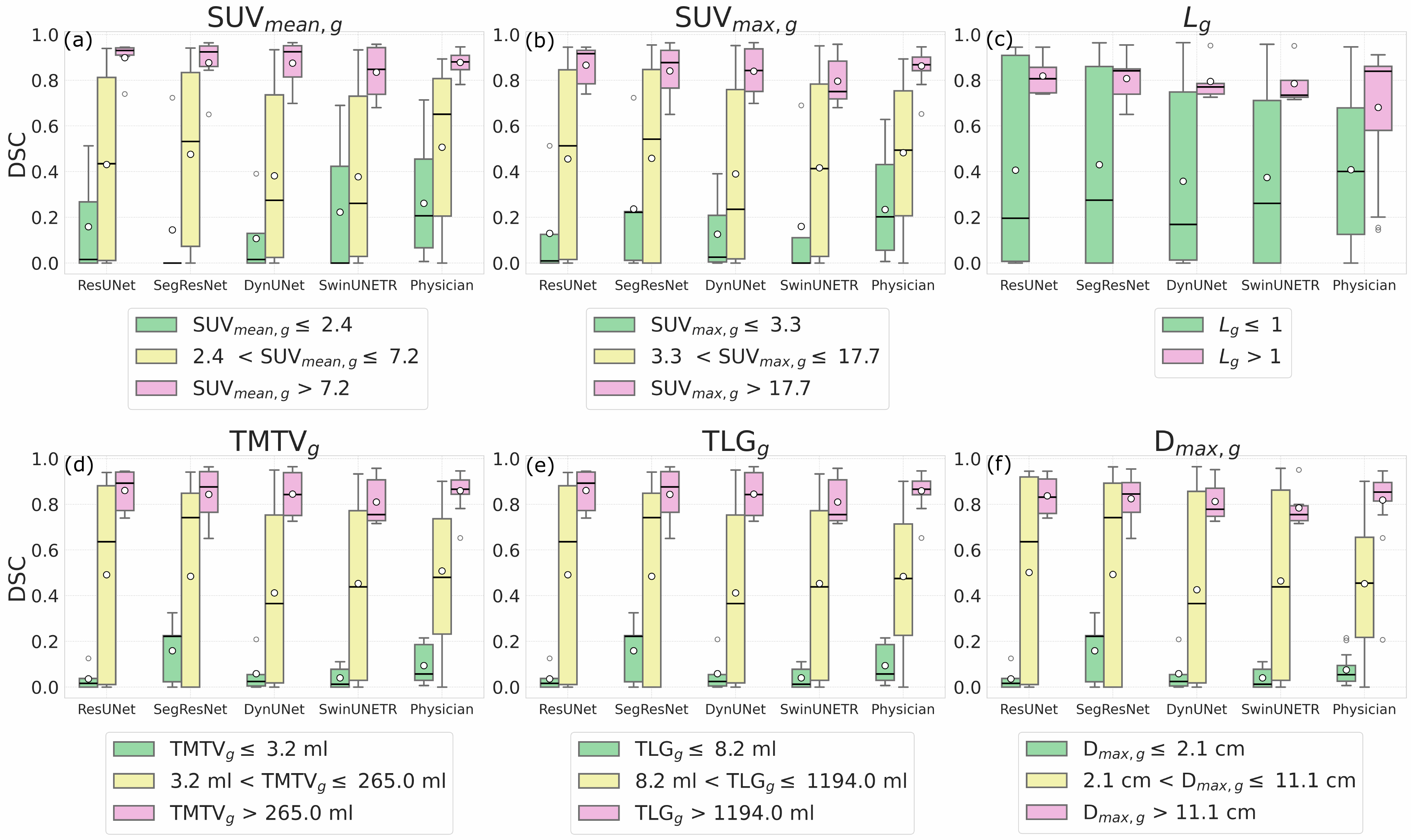}
\caption{DSC performance distributions of networks on the PMBCL-BCCV test set, as well as Physician 3 on the 60 cases of the set used for assessing intra-observer agreement. Analysis is performed across three different categories: PMBCL-BCCV test set (i) lesion measure $\leq$ 20\%tile, (ii)  20\%tile < lesion measure $\leq$ 75\%tile, (iii) lesion measure > 75\%tile. The DSC performance of physician mirrors that of networks. The values of the 20\%tile and 75\%tile thresholds of  lesion measures are listed in the boxes below each plot.}
\label{fig:lesion_measures_segregated_dsc_metrics_physician_vs_networks}
\end{figure*}

To perform intra-observer variability analysis, 60 cases from the entire PMBCL-BCCV cohort (encompassing train, valid, and test sets) were re-segmented by Physician 3. To eliminate bias, the selection of these cases was randomized, ensuring no preference in the selection of specific cases was given during the re-segmentation process. The mean DSC between original and new segmentations by Physician 3 over the 60 cases was $0.50 \pm 0.33$. The mean was comparable to the PMBCL-BCCV test set mean DSC performance of ResUNet ($0.49 \pm 0.42$) and SegResNet ($0.51 \pm 0.41$) (see \Cref{tab:final_segmentation_results}).

Additionally, we assessed the reproducibility of lesion measures between the original and new segmentations via equivalence testing at $\alpha = 0.05$, as shown in \Cref{fig:equivalence_testing_intraobs_analysis_segregated_in_three_categories}. The analyses was performed by grouping the cases into three categories: where GT lesion measure $\leq$ 20\%tile, 20-75\%tile, and >75\%tile. These 20\%tile and 75\%tile thresholds were chosen on the PMBCL-BCCV test set for all lesion measures except for the lesion measure $L$, where these thresholds were chosen on these 60 cases itself, since the 20\%tile and 75\%tile values for $L$ on PMBCL-BCCV test set were both 1 (see \Cref{fig:lesion_measures_segregated_dsc_metrics_physician_vs_networks} (c)). Physician 3's annotations were consistent enough to produce equivalent (reproducible) SUV$_\text{mean}$ on all three categories. For SUV$_\text{max}$, the physician was reproducible for all categories except $\leq 20\%$tile SUV$_\text{max}$. This makes practical sense since physician might be prone to missing out SUV$_\text{max}$ voxels in cases when they are very small. Similarly, for the TMTV and TLG measures, the physician was consistent over all categories except the $\leq 20\%$tile category. This shows that it is hard even for the physician to consistently segment lesions that are small and metabolically less active. Finally, the $L$ and D$_\text{max}$ measures were not reproducible for any categories.

\subsubsection{Networks vs. physician: segmentation performance}
\label{networks_vs_physician_segmentation_performance}

To compare the segmentation performance of physician to that of networks, we performed an analysis similar to \Cref{fig:lesion_measures_segregated_dsc_metrics_segresnet} but for the PMBCL-BCCV test set for all networks and for the subset of the entire PMBCL-BCCV set (with 60 cases) which were used to assess intra-observer variability by Physician 3. We chose the same values for the 20\%tile and 75\%tile thresholds for GT lesion measures on the 60 cases as was computed on the PMBCL-BCCV test set. To assess physician's performance, the DSC was computed between the original and new annotations (i.e.~physician DSC performance is also a measure of intra-observer agreement). The results are presented in \Cref{fig:lesion_measures_segregated_dsc_metrics_physician_vs_networks}, showing that the performance of physician within different lesion measure categories mirrors that of networks. That is, the categories on which the lesion measures such as $\text{SUV}_\text{mean}$, $\text{SUV}_\text{max}$, TMTV, and TLG had the lowest values where the networks had the most segmentation errors, the physician was also inconsistent on those categories and vice-versa.

Additionally, the physician's performance distribution on the DSC metrics, as shown in \Cref{fig:lesion_measures_segregated_dsc_metrics_physician_vs_networks} was used as a measure for the intra-observer agreement for segmentation. Specifically, on the TMTV measure categories, the physician had a mean DSC of $0.09\pm0.08$, $0.51\pm0.27$, and $0.86\pm0.07$ in the categories, TMTV$\leq$3.2 ml (<20\%tile), 3.2 ml$<$TMTV$\leq$265.0 ml (20-75\%tile), and TMTV>265.0 ml (>75\%tile) categories, respectively. The DSC distribution for other lesion measures have been presented in \Cref{fig:lesion_measures_segregated_dsc_metrics_physician_vs_networks} (a)-(c), (e)-(f).

\subsubsection{Networks vs. physician: detection performance}
\label{networks_vs_physician_detection_performance}
\begin{figure}[h!]
\centering
\includegraphics[width=0.84\linewidth]{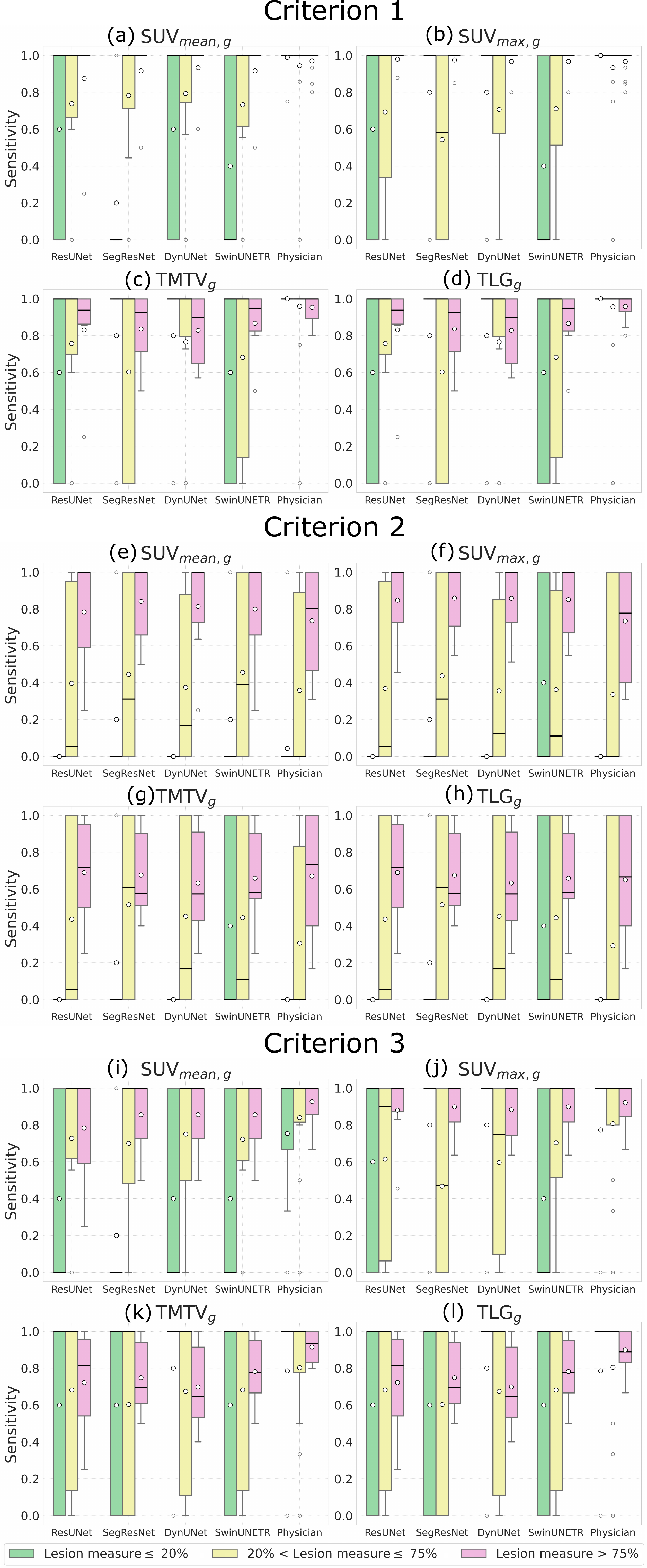}
\caption{Comparison of networks and physician's (Physician 3) performance (intra-observer agreement) using detection sensitivity within different categories of various lesion measures. Top, middle, and bottom panels show sensitivity distribution under Criteria 1, 2, and 3, respectively.}
\label{fig:lesion_measures_segregated_sensitivityC1C2C3_metrics_physician_vs_networks_vertical}
\end{figure}

To compare the detection performance of physician to that of networks on the three detection criteria, we again performed an analysis similar to above but for the detection sensitivity metric instead of DSC. For the detection analyses, we used the same 20\%tile and 75\%tile thresholds on GT lesion measures from the PMBCL-BCCV test set as in \Cref{fig:lesion_measures_segregated_dsc_metrics_physician_vs_networks}. The results are presented in \Cref{fig:lesion_measures_segregated_sensitivityC1C2C3_metrics_physician_vs_networks_vertical}.

On Criterion 1, the physician demonstrated consistent performance on the  $\text{SUV}_\text{mean}$ and $\text{SUV}_\text{max}$ measures across all categories. However, for TMTV and TLG measures, slight inconsistencies were observed, particularly in the higher categories (TMTV > 265.0 ml and TLG > 1194.0 ml, each > 75\%tile), as shown by the broader range of sensitivities in \Cref{fig:lesion_measures_segregated_sensitivityC1C2C3_metrics_physician_vs_networks_vertical} (c)-(d). This indicates that, while the physician consistently segmented at least one voxel within the ground truth lesion, larger TMTV or TLG values (associated with numerous small lesions or fewer large lesions) introduced some variability. This variability suggests that in cases with a high number of lesions, the physician may occasionally miss some lesions, leading to reduced sensitivity under Criterion 1. This pattern is understandable, as segmenting numerous lesions can increase the likelihood of omission.

On Criterion 2, the physician's performance largely mirrored that of networks. All networks and physician had sensitivities distributed around 0 on all lesion measures for the lower 20\%tile. This shows that both networks and physician could not segment lesions with IoU > 50\% whenever the lesions were faint (lower $\text{SUV}_\text{mean}$ or $\text{SUV}_\text{max}$) or small (lower TMTV or TLG). Surprisingly, for the higher category (>75\%tile) for all lesions measures, all networks had a narrower sensitivity distributions as compared to that of physician showing that physician was more inconsistent (thereby less reproducible) on such cases. On the other hand, the performance of networks were largely consistent with the physician for middle category (20\%-75\%tile of lesion measures) as shown by the broad distribution of sensitivities lying between 0 and 1 (\Cref{fig:lesion_measures_segregated_sensitivityC1C2C3_metrics_physician_vs_networks_vertical}).

On Criterion 3, we observed that the physician was highly consistent with the segmentation of SUV$_\text{max}$ voxel within lesions as compared to networks even for the lower $<20\%$tile category over all lesion measures (\Cref{fig:lesion_measures_segregated_sensitivityC1C2C3_metrics_physician_vs_networks_vertical} (i)-(l)). This shows that, unlike networks (see \Cref{fig:lesion_measures_segregated_dsc_metrics_segresnet,fig:dsc_vs_gt_biomarkers_combined}), even for the small and faint lesions, the physician was always good at locating the lesion (by accurately segmenting the SUV$_\text{max}$ voxel, as shown under Criterion 3), although not consistently segmenting the boundary of lesions (as shown under Criterion 2).

To highlight the difference between the detection sensitivity under Criteria 2 (commonly used in computer vision) and 3 (our proposed criterion), we show the various mean detection sensitivity values for physician and networks in \Cref{tab:detection_sensitivity_tmtv_categories} for different categories of TMTV. The networks consistently obtained higher sensitivity under Criterion 3 as compared to Criterion 2 for all categories of TMTV showing that across all categories of tumor burden, the sensitivity metric under Criterion 3 is superior (and hence more informative) of networks ability to localize lesions. However, physician demonstrated superior detection performance on Criterion 3 across all TMTV categories compared to the networks, underscoring their exceptional ability to identify lesion SUV$_\text{max}$ voxels. This advantage could be attributed to their extensive clinical training and years of experience reading PET scans.

\begin{table}[]
\centering
\resizebox{\columnwidth}{!}{%
\begin{tabular}{@{}ccccc@{}}
\toprule
\multirow{2}{*}{\textbf{\begin{tabular}[c]{@{}c@{}}TMTV \\ category\end{tabular}}} &
  \multirow{2}{*}{\textbf{Methods}} &
  \multicolumn{3}{c}{\textbf{Detection sensitivity}} \\ \cmidrule(l){3-5} 
                                       &           & \textbf{Criterion 1} & \textbf{Criterion 2} & \textbf{Criterion 3} \\ \midrule
\multirow{5}{*}{$\leq$ 3.2 ml}         & ResUNet   & 0.6 $\pm$ 0.55           & 0.0 $\pm$ 0.0            & 0.6 $\pm$ 0.55           \\
                                       & SegResNet & 0.8 $\pm$ 0.45           & 0.2 $\pm$ 0.45           & 0.6 $\pm$ 0.55           \\
                                       & DynUNet   & 0.8 $\pm$ 0.45           & 0.0 $\pm$ 0.0            & 0.8 $\pm$ 0.45           \\
                                       & SwinUNETR & 0.6 $\pm$ 0.55           & 0.4 $\pm$ 0.55           & 0.6 $\pm$ 0.55           \\
                                       & Physician & 1.0 $\pm$ 0.0            & 0.0 $\pm$ 0.0            & 0.79 $\pm$ 0.43          \\ \midrule
\multirow{5}{*}{3.2 ml - 265.0 ml}     & ResUNet   & 0.76 $\pm$ 0.42          & 0.44 $\pm$ 0.51          & 0.68 $\pm$ 0.46          \\
                                       & SegResNet & 0.6 $\pm$ 0.49           & 0.52 $\pm$ 0.51          & 0.6 $\pm$ 0.49           \\
                                       & DynUNet   & 0.77 $\pm$ 0.42          & 0.45 $\pm$ 0.5           & 0.67 $\pm$ 0.47          \\
                                       & SwinUNETR & 0.68 $\pm$ 0.46          & 0.44 $\pm$ 0.5           & 0.68 $\pm$ 0.46          \\
                                       & Physician & 0.96 $\pm$ 0.18          & 0.31 $\pm$ 0.44          & 0.8 $\pm$ 0.36           \\ \midrule
\multirow{5}{*}{\textgreater 265.0 ml} & ResUNet   & 0.83 $\pm$ 0.29          & 0.69 $\pm$ 0.3           & 0.72 $\pm$ 0.31          \\
                                       & SegResNet & 0.84 $\pm$ 0.21          & 0.68 $\pm$ 0.26          & 0.75 $\pm$ 0.21          \\
                                       & DynUNet   & 0.83 $\pm$ 0.2           & 0.63 $\pm$ 0.31          & 0.7 $\pm$ 0.25           \\
                                       & SwinUNETR & 0.87 $\pm$ 0.2           & 0.66 $\pm$ 0.29          & 0.78 $\pm$ 0.2           \\
                                       & Physician & 0.95 $\pm$ 0.07          & 0.67 $\pm$ 0.31          & 0.92 $\pm$ 0.09          \\ \bottomrule
\end{tabular}%
}
\caption{Detection performance of networks and physician under the three detection criteria for different categories of TMTV: TMTV<3.2 ml (<20\%tile), 3.2 ml<TMTV $\leq$ 265.0 ml (20\%tile - 75\%tile), and TMTV>265.0 ml (>75\%tile). All networks consistently show superior performance on Criterion 3 as compared to Criterion 2 highlighting its potential clinical utility.}
\label{tab:detection_sensitivity_tmtv_categories}
\end{table}

\subsection{Inter-observer variability}
\label{subsec:inter_observer_variability}

35 cases were randomly selected from the DLBCL-BCCV set that were segmented by Physicians 2, which were used to quantify inter-observer variability between Physicians 1 and 2. The mean DSC between the two physicians was $0.78 \pm 0.13$, while the mean DSC with respect to the STAPLE-generated consensus was $0.87 \pm 0.13$ and $0.89 \pm 0.08$ for Physicians 1 and 2 respectively.  Moreover, the ICC between the two physicians on SUV$_\text{mean}$, SUV$_\text{max}$, $L$, TMTV, TLG, $D_\text{max}$ were found to be 0.99, 1.00, 0.98, 0.99, 1.00, 0.93 respectively, showing considerably high agreement in the estimation of lesion measures.

\section{Discussion}
\label{sec:discussion}

In this work, we trained and evaluated four distinct neural networks to automate the segmentation of lymphoma lesions from PET/CT datasets sourced from three different cohorts. To assess models performance, we conducted comprehensive evaluations on internal test set originating from these three cohorts and showed that SegResNet and ResUNet outperformed DynUNet and SwinUNETR on the DSC (mean and median) and median FPV metrics, while SwinUNETR had the best median FNV. Additionally, we extended our analysis to encompass an external out-of-distribution testing phase on a sizable public lymphoma PET/CT dataset. On this external test set as well, SegResNet emerged as the top performer in terms of DSC and FPV metrics, underscoring its robustness and effectiveness, while ResUNet was the best on FNV.\\

\textbf{Network architectures and training protocols.} SegResNet and ResUNet, with their convolutional architectures, are highly effective at capturing local features and spatial hierarchies, crucial for lesion segmentation. DynUNet’s adaptable structure could potentially offer improved performance by tuning to the data, but it may also perform suboptimally if the dynamic adjustments are not optimal for the specific task. Moreover, transformer-based SwinUNETR learns global representations providing understanding of global context in the data, which is beneficial for complex segmentations but might be less effective for tasks where local details are more critical (such as small lesion segmentation) or when the labeled data is scarce.

Additionally, it is important to highlight that SegResNet and ResUNet were trained on patches of larger sizes, specifically with $N=224$ and $N=192$, respectively (see detailed training methodology and hyperparameter choices in \Cref{app:network_architectures_and_training_setups}), while DynUNet and SwinUNETR were trained using relatively smaller sized patches, namely $N=160$ and $N=128$, respectively. Utilizing larger patch sizes during training allows the neural networks to capture a more extensive contextual understanding of the data, thereby enhancing its performance in segmentation tasks \cite{Isensee_2020}. This observation aligns with our results, where the superior performance of SegResNet and ResUNet can be attributed to their exposure to larger patch sizes during training. Moreover, larger batch sizes enable robust training by accurately estimating the gradients \cite{Isensee_2020}, but with our chosen training patch sizes, we could not train SegResNet, DynUNet and SwinUNETR with $n_b > 1$ due to memory limitations. Hence, for a fair comparison between networks, all networks were trained with $n_b=1$. It is worth noting that our inability to train DynUNet and SwinUNETR on larger patch and mini-batch sizes was primarily due to computational resource limitations, which presents an avenue for future research, where training these networks on larger patches/batch sizes could potentially yield further performance improvements. \\

\textbf{Reproducibility of lesion measures by networks.} In alignment with the RELAINCE guidelines for validation of AI-based clinical decision support tools in nuclear medicine \cite{relaince_paper}, we structured our study to include both proof-of-concept and task-based evaluations to comprehensively assess the performance of deep neural networks for lymphoma lesion segmentation from PET/CT images. The RELAINCE guidelines emphasizes that while proof-of-concept evaluations, based on conventional metrics like DSC are essential to establish the technical feasibility of AI models, they do not sufficiently capture clinical utility. Thus, following RELAINCE's guidelines, we performed task-based evaluations that quantify model performance through clinically-relevant lesion measures such as TMTV, TLG, and SUV metrics (along with detection sensitivity under various detection criteria), enabling the assessment of clinical reproducibility and reliability of these models in real-world settings.

We assessed the reproducibility of lesion measures and found that on internal testing, SUV$_\text{mean}$ was reproducible by all networks except ResUNet, TLG was reproducible by all networks, while no other measure was reproducible by any network. In external testing, the reproducibility was further limited, with SUV$_\text{mean}$ reproducible by all networks except ResUNet, and TLG reproducible by only DynUNet, while none of the other measures were reproducible by any network. Although achieving high DSC scores, these networks demonstrated limitations in accurately replicating clinically significant metrics. Under this framework, the observed lack of equivalence for lesion measures, especially, SUV$_\text{max}$ and TMTV (even within internal testing) also highlights intrinsic limitations of these measures rather than a failure of the segmentation models. SUV$_\text{max}$ is highly sensitive to voxel-level localization and partial-volume effects, while TMTV aggregates segmentation errors across lesions, amplifying small boundary differences. Clinically, this suggests that although automated methods may reliably support lesion detection and qualitative assessment of disease burden, derived quantitative measures such as SUV$_\text{max}$ and TMTV should be interpreted with caution when used for precise numerical decision-making, including response assessment or prognostic stratification. Importantly, comparable variability is also observed among expert human readers, underscoring the need to evaluate automated performance in the context of expected clinical and observer variability rather than overlap metrics alone.

Furthermore, we quantified the errors in the estimation of lesion measures using modified Bland-Altman plots and found that in certain ranges of GT lesion measure values, the predicted lesion measures were highly under or overestimated even by networks with high median DSC on the combined internal and external test sets (\Cref{fig:modified_bland_altman_plot}). The networks generally made significant errors in prediction when GT lesion measures were very small. We also showed that, in general, on a set of images with larger patient level lesion $\text{SUV}_\text{mean}$, $\text{SUV}_\text{mean}$, TMTV, and TLG, a network is able to predict with a higher median DSC, although for very high values of these lesion measures, the performance generally plateaus. 

The modified Bland-Altman plots revealed that large outliers in lesion measure estimation were driven by a relatively smaller subset of challenging cases with consistent imaging characteristics. Overestimation outliers - particularly for SUV$_\text{max}$ - were most commonly associated with false positive predictions in regions of physiological high FDG uptake, including the brain, myocardium, kidneys, urinary bladder, bowel, and occasionally brown fat or inflammatory uptake. When such false positives were assigned high SUV values, even a single voxel could substantially inflate SUV$_\text{max}$ and, in some cases, propagate to volumetric measures such as TMTV and TLG. Conversely, large underestimation errors were typically observed in cases with small, faint, or highly heterogeneous lesions, where partial-volume effects and low lesion-to-background contrast made accurate boundary delineation difficult. Quantitatively, these cases were characterized by elevated FPV and were more prevalent in the lower quantiles of lesion SUV and TMTV, consistent with the lesion-measure-stratified analyses (\Cref{fig:lesion_measures_segregated_dsc_metrics_segresnet,fig:lesion_measures_segregated_dsc_metrics_physician_vs_networks}) and observer variability results (\Cref{fig:equivalence_testing_intraobs_analysis_segregated_in_three_categories,fig:lesion_measures_segregated_dsc_metrics_physician_vs_networks}). Importantly, similar failure modes were also observed in expert annotations, indicating that these outliers reflect intrinsic challenges of PET/CT interpretation rather than isolated model-specific errors.

On the other hand, the DSC performance was not much affected by the number of lesions, while for a set of images with higher $\text{D}_\text{max}$, the performance generally decreased for all networks (\Cref{fig:dsc_vs_gt_biomarkers_combined}). Accessing diverse PET/CT datasets for training and testing models is challenging due to the private ownership of healthcare data. To enhance model interpretability, researchers must examine how their model performance varies with dataset characteristics. By exploring correlations between model performance and image/lesion measures, it is possible to uncover both strengths and limitations of these models \cite{tumor_feature_performance_dependence}. 

As we have shown in our work, a high overlap-based segmentation performance does not necessarily imply accurate reproduction of clinically relevant PET-derived lesion measures. Metrics such as DSC primarily quantify volumetric overlap and are relatively insensitive to systematic boundary biases or localized intensity-weighted errors. As a result, networks such as ResUNet can achieve high DSC by accurately capturing the bulk of large lesions while still exhibiting small but systematic boundary deviations. Although these deviations have minimal impact on DSC, they can substantially affect intensity-weighted measures such as SUV$_\text{mean}$, particularly in heterogeneous or low-contrast lesions. These effects are further amplified under domain shift in external datasets, where differences in scanner characteristics, reconstruction protocols, and uptake distributions alter intensity profiles without substantially degrading spatial overlap.

Moreover, all networks in this study were trained using Dice loss and optimized for voxel-wise segmentation accuracy rather than for the preservation of quantitative PET-derived biomarkers. This follows common practice in medical image segmentation, where training objectives are designed to optimize spatial agreement rather than downstream quantitative readouts. Consequently, none of the architectures were explicitly encouraged during training to maintain fidelity of intensity-weighted clinical measures, which likely contributes to the observed degradation in SUV-based reproducibility, particularly in out-of-distribution settings. We also observed that the transformer-based SwinUNETR exhibited more stable performance trends across internal and external datasets for some lesion measures, likely reflecting its ability to model longer-range spatial context and global dependencies. However, this did not consistently translate into superior reproducibility of all clinical metrics. Under the controlled training setup used in this study, the observed discrepancies appear to be driven primarily by training objectives and data characteristics, with architectural effects playing a secondary role. A systematic evaluation combining architectural choices with task-aware loss functions explicitly targeting clinically-relevant PET-derived lesion measures represents an important direction for future work.\\

\textbf{Detection performance by networks.} In addition to evaluating segmentation performance, we introduced three detection criteria (Criterion 1, 2, and 3) to assess the networks' performance on a per-lesion basis. Unlike segmentation assessment, which emphasizes voxel-level accuracy and boundary delineation, these criteria focus on how well the networks identify and locate lesions within the images. Criterion 1 evaluates the networks' ability to detect the presence of lesions, as it directly influences whether a potential health concern is identified or missed. Detecting even a single voxel of a lesion could trigger further investigation. Criterion 2 assesses lesion count and localization, and can significantly impact therapeutic decisions and treatment planning. Moreover, Criterion 3 adds an additional layer of clinical relevance by segmenting lesions based on their metabolic characteristics ($\text{SUV}_\text{max}$). Using these detection criteria, we assessed the sensitivities and FP detections for all networks and showed that depending on the detection criteria, a network can have very high sensitivity even when the DSC performance was low. Given these different detection criteria, a trained model can be chosen based on specific clinical use cases. For example, certain use cases might require the physician to be able to detect all lesions without being overly cautious about the exact lesion boundary, while other use cases might be looking for more robust boundary delineations.\\

\textbf{Intra- and inter-observer variability analysis and physician vs. networks performances.} We assessed physician intra-observer variability in segmenting cases with a range of values for lesion measures, noting challenges in the consistent segmentation of cases from the $\leq$20\%tile categories, i.e.~the cases with low SUV$_\text{mean}$, SUV$_\text{max}$, TMTV, and TLG. In lymphoma lesion segmentation, cases can vary in difficulty due to factors like size, shape, and location of lesions, or image quality. By identifying which cases are consistently difficult for even an experienced physician to segment, we gained insights into the complexities and nuances of the segmentation task. The DSC metric computed to assess the intra-observer agreement between the physician's original and new segmentations (on the PMBCL cohort) was comparable to the networks performance on the test set from the same cohort. Moreover, the segmentation performance by physician on cases categorized by different ranges of lesion measure values largely mirrored the performance by networks in terms of distribution of DSC over different categories. 

Similarly, we also compared the detection performance between physician and networks using the sensitivity metric under the three detection criteria. Under Criterion 1, the physician was largely adept at segmenting at least one voxel of the ground truth lesion across all categories of lesion measures, while the networks usually struggled with segmenting lesions with lower values of lesion measures. We attribute this difference to the clinical training of physicians that help them identify even small lesions while the network usually fails on those cases due to various reasons such as limited spatial resolution and pooling effects, limited receptive field sensitivity, reduced lesion-to-background constrast, and insufficient training data with small lesions. Under Criterion 2, the the physician's performance mirrored that of networks in terms of the distribution of sensitivities over all three categories, where both networks and physician had poor performance on $\leq$20\%tile category of lesion measures. This shows that, although the physician was adept at detecting at least one voxel of even small/faint lesions, neither physician nor networks were consistent with segmenting at least 50\% volume of the lesion accurately in cases with small/faint lesions. This observation further highlights the difficulty of PET lesion segmentation task, where the images suffer from low spatial resolution, blurring, high levels of noise, partial volume effects, and weak and diffuse boundaries between lesion and background. Finally, under Criterion 3, we observed that all networks had superior sensitivities as compared to their sensitivities under Criterion 2. 

We interviewed several physicians to gather their perspectives on lesion detection. They expressed concerns regarding relying on Criterion 2 to evaluate lesion detection. Lesion boundaries are often not well-defined in PET imaging due to a relatively low spatial resolution, low signal-to-background ratio, partial volume effects, etc. This is reflected in the detection performance of physician under Criterion 2 in \Cref{tab:detection_sensitivity_tmtv_categories}. On the other hand, Criterion 1 fails to effectively capture the model's ability to detect the most metabolically active regions within lesions, which is clinically crucial for assessing tumor aggressiveness. This feedback motivated us to propose Criterion 3. In this work, we show the superiority of our proposed Criterion 3 over existing Criteria 1 and 2 for evaluating networks lesion localization performance. By introducing Criterion 3, we aim to emphasize that in clinical settings, it may be sufficient to identify all lesions using their SUV$_\text{max}$ markers \cite{hirata2021preliminary}, rather than precisely delineating lesion volumes, to effectively trigger a diagnosis. For small and faint lesions, the GT boundary delineation can itself be challenging given low signal-to-background ratio, partial volume effects, etc., as was observed in the detection performance of physician under Criterion 2 for <20\%tile category of lesion measures.

As determined by DSC, the inter-observer agreement was $\sim$28\% higher than the intra-observer agreement. We attribute this stark difference to the fact that the inter-observer analysis was performed on a subset of DLBCL-BCCV cohort while the intra-observer analysis was performed on a subset of PMBCL-BCCV cohort. As shown in \Cref{fig:distribution_of_lesion_measures}, the PMBCL-BCCV was a much richer dataset with a larger variation in TMTV and TLG (with the distribution of $\leq$20\%tile TMTV/TLG category of PMBCL-BCCV lying around values smaller than the corresponding distribution for $\leq$20\%tile TMTV/TLG category of DLBCL-BCCV) along with lower mean SUV$_\text{mean}$ and SUV$_\text{max}$ as compared to DLBCL-BCCV. Hence, we conclude that the PMBCL-BCCV cohort was also more challenging to segment, making it harder for the physician to consistently generate GT (especially on cases with lower values of TMTV/TLG). Since observer variability is a fundamental limit to the measurable model performance, this also reflected in the worse performance of model on these cases.

Importantly, it is worth noting that image interpretation itself is not wholly objective: cognitive heuristics systematically bias what physicians perceive and report, introducing additional reader-to-reader variability \cite{busby2018bias}. Acknowledging this human factor underscores the value of formally quantifying intra- and inter-observer agreement and cautions against over-interpreting performance differences that may lie within the noise imposed by such biases. \\

\textbf{Segmentation of small and faint lesions.} Although the methods used achieve consistently high performance on large, high-uptake lesions, a systematic analysis showed a pronounced drop in DSC and detection sensitivity for the $\leq$20\%tile strata of SUV$_\text{mean}$, SUV$_\text{max}$, TMTV and TLG, i.e. very small or metabolically faint foci (as shown in \Cref{fig:lesion_measures_segregated_dsc_metrics_segresnet,fig:dsc_vs_gt_biomarkers_combined,fig:lesion_measures_segregated_dsc_metrics_physician_vs_networks}) for both networks and physician. This weakness reflects fundamental imaging constraints (partial-volume blurring, low signal-to-noise) and is exacerbated by class-imbalance in the training data and the absence of multi-reader consensus masks. For future work, we plan to (i) enrich the dataset with scans that purposely oversample indolent or residual disease and to synthesize micro-lesions for data augmentation \cite{de2025characterization}; (ii) investigate lesion-aware augmentation \cite{lama2024lama}, cascaded hotspot-guided networks \cite{su2023attention} and multi-scale transformer–CNN hybrids that operate at higher native resolution \cite{li2023mctnet}; (iii) exploit PET-CT fusion strategies such as multimodal spatial attention \cite{petctmultimodal1} and mirror UNet \cite{petctmultimodal2} to capture the subtle CT cues that accompany PET-occult disease. These directions, together with active-learning loops \cite{budd2021survey} that solicit expert feedback on low-confidence outputs, are expected to close the performance gap for small and faint lesions and to broaden the clinical applicability of automated lymphoma segmentation.\\

\textbf{Clinical insights for multi-modality model training.} On a closer look, a considerable fraction of cases within the <20\%tile category in \Cref{fig:lesion_measures_segregated_dsc_metrics_physician_vs_networks} (a)-(b),(d)-(e) from the PMBCL-BCCV test cohort were related to residual disease in the thymus gland with negligibly low uptake (and very small TMTV) on PET but noticeable structural abnormality on CT. Hence, unlike the cases presenting lesions with high SUV (where PET is the modality containing the primary source of signal for training models), for this category, the primary source of signal was contained in CT. Since these cases were still a very small fraction of the overall PMBCL-BCCV training set, the models failed to learn CT-based representations when the signal from PET was vanishingly small. To mitigate this, techniques such as multimodal spatial attention \cite{petctmultimodal1} and mirror UNet \cite{petctmultimodal2} that aim to better utilize the complementary information from multimodal data will be explored in future work.\\    

\textbf{Histology-agnostic robustness.} Although the present study was conducted mainly for high-grade B-cell lymphomas (DLBCL and PMBCL), its histological scope remains narrow. Indolent B-cell lymphomas (follicular, marginal-zone, mantle-cell) \cite{smyth2023management}, mature T-cell lymphomas \cite{savage2023introduction}, and Hodgkin lymphoma \cite{momotow2021hodgkin} differ markedly in metabolic profile, lesion distribution, and treatment strategy -- factors that can alter tracer uptake patterns and, by extension, model behavior. To ensure histology-agnostic robustness of our models, we have now begun curating additional PET/CT datasets that include these additional subtypes and are actively training models on them as part of our future work. We believe these steps together will extend our models applicability beyond the high-grade B-cell setting evaluated in this work.\\

\textbf{Need for a rigorous multi-physician ground truth segmentation protocol.} Another limitation of our work is that the GT annotation used for training the models were defined by a single physician (Physicians 1, 3, and 5) for each of the internal cohorts, hence no case benefitted from annotations from multiple experts. To improve GT consistency, a well-defined and robust segmentation protocol that engages multiple experts to delineate cases independently and also with feedback from one another is essential. The independent annotations from multiple experts can also be combined via STAPLE to generate GT consensus, which can then be used to train models, although such delineations are often hard to obtain in practical scenario. Alternatively, deep learning methods like D-LEMA \cite{mirikharaji2021d} can be explored that train the network by utilizing annotations from multiple experts without the need for generating consensus.

\section{Conclusion}
\label{sec:conclusion}
In this work, we assessed multiple neural network architectures for automating lymphoma lesion segmentation in PET/CT images across multiple datasets. We  examined the reproducibility of lesion measures, revealing differences among networks, highlighting their suitability for specific clinical uses. Additionally, we used different lesion detection criteria to assess network performance at a per-lesion level, emphasizing their clinical relevance. Lastly, we performed observer variability analyses to emphasize the challenges related to ground truth consistency. Our work provides insights into deep learning's potentials and limitations in lymphoma lesion segmentation, emphasizing the need for standardized annotation practices to enhance research validity and clinical applicability.

\bibliographystyle{unsrtnat}
\bibliography{main_bibliography}

\section*{Statements and declarations}
\renewcommand{\thesubsection}{S\arabic{subsection}}
\subsection{Funding}
This work was supported by Mitacs Accelerate Grant IT28444 (jointly via Mitacs Inc., Canada and Microsoft AI for Good Lab, Redmond, USA), as well as Canadian Institutes of Health Research (CIHR) Grant PIBH-GR018169. Computational resources were provided by Microsoft AI for Good Lab, Redmond, USA.
\subsection{Competing interests}
Nothing to disclose.
\subsection{Ethics approval}
\label{ethical_statements}
We used four data cohorts in this study consisting of PET/CT images of human patients presenting different lymphoma phenotypes. For DLBCL-BCCV and PMBCL-BCCV cohorts, the ethics approval was granted by the UBC BC Cancer Research Ethics Board (REB) (REB Numbers: H19-01866 and H19-01611 respectively) on 30 Oct 2019 and 1 Aug 2019 respectively. For DLBCL-SMHS cohort, the approval was granted by St. Mary's Hospital, Seoul (REB Number: KC11EISI0293) on 2 May 2011. Due to the retrospective nature of these datasets, patient consent was waived for these three cohorts. These cohorts also comprise of datasets that are privately-owned by the respective hospitals. Lastly, the AutoPET lymphoma dataset was a publicly hosted dataset associated with the AutoPET challenge 2023 (MICCAI) and was acquired from The Cancer Imaging Archive. 
\subsection{Author contributions}
The study design, model training, and result analysis were conducted by Shadab Ahamed, with technical inputs from Yixi Xu through regular discussions. Sara Kurkowska, Claire Gowdy, Joo H. O, Ingrid Bloise, Don Wilson, and Patrick Martineau were responsible for data annotation. The manuscript benefited from clinical insights provided by Sara Kurkowska, Claire Gowdy, and Ingrid Bloise. Fran\c{c}ois B\'{e}nard, Fereshteh Yousefirizi, Carlos F. Uribe and Arman Rahmim were involved in approval and acquisition of datasets. The manuscript was drafted by Shadab Ahamed, and all authors commented on previous versions of the manuscript with key contributions from Yixi Xu, Rahul Dodhia, Juan M. Lavista, William B. Week, and Arman Rahmim. All authors approved and consented to the submission of the final manuscript.  

\clearpage
\appendix
\section*{Appendix}
\renewcommand{\thetable}{A\arabic{table}}
\setcounter{table}{0}
\renewcommand{\thesubsection}{\Alph{subsection}}
\renewcommand{\thefigure}{A\arabic{figure}}
\setcounter{figure}{0}
\subsection{Network architectures and training tools}
\label{app:network_architectures_and_training_setups}
All the networks were implemented using \texttt{MONAI} \cite{monai_paper} and PyTorch, and experiments were conducted on a Microsoft Azure virtual machine  with Ubuntu 16.04, 12 CPU cores (224 GiB RAM) and 2 NVIDIA Tesla V100 GPU (16 GiB RAM each). The architectural details for the four deep neural networks used in this work are given below:
\begin{enumerate}[label=(\alph*)]
\item \texttt{ResUNet}: The ResUNet architecture based on \cite{residual_unet} consisted of 6 layers of encoder and decoder (with residual blocks) paths with skip-connections. The data in the encoder was downsampled using strided convolutions, while the decoder unsampled using transpose strided convolutions. The number of channels in the encoder part from the top-most layer to the bottleneck were 16, 32, 64, 128, 256, and 512 and PReLU was used as activation function. This network consisted of 19,289,401 trainable parameters.
\item \texttt{SegResNet}: The SegResNet architecture based on \cite{segresnet} consisted of encoder part which used ResNet \cite{resnet} blocks and included 4 stages of 1, 2, 2, 4 blocks, respectively. The upsampling blocks consisted of a series of residual blocks followed by non-trainable operation that carried out the upsampling via trilinear interpolation. This network used ReLU as activation functions and consisted of 4,701,346 trainable parameters. 
\item \texttt{DynUNet}: Dynamic UNet or DynUNet is a reimplementation of the \texttt{3D-fullres} implementation of nn-UNet \cite{isensee2018nnunet}. For this work, we implemented a DynUNet with 6 layers of encoder and decoder blocks with skip-connections. The encoder blocks consisted of convolutional layers intercepted by Leaky ReLU activations and the upsampling was performed via transpose convolution operations. This network consisted of 31,190,274 trainable parameters. 
\item \texttt{SwinUNETR}: The SwinUNETR architecture based on \cite{swinunetr} consisted of a swin transformer encoder for extracting features at 5 different resolutions by utilizing shifted windows for computing self-attention and was connected to a CNN-based decoder at each resolution via skip connections. For this network, we used sigmoid activation function in the encoder path and Leaky ReLU in the decoder path and the network consisted of 4,078,496 trainable parameters.
\end{enumerate}

\subsection{Training and inference methodology}
\label{app:training_details}
\subsubsection{Preprocessing and augmentations}
\label{subsubsec:data_preprocessing_and_augmentations}
The high-resolution CT images in Hounsfield unit (HU) were downsampled to the coordinates of their corresponding PET/mask images. The PET intensity values in units of Bq/ml were decay-corrected and converted to SUV. 

During training, we employed a set of non-randomized and randomized transforms to augment the input to the network. The non-randomized transforms included (i) clipping CT intensities in [-154, 325] HU followed by min-max normalization, (ii) cropping the region outside the body in PET, CT, and mask images using a  3D bounding box, and (iii) resampling the images to an isotropic voxel spacing of (2.0 mm, 2.0 mm, 2.0 mm) via bilinear interpolation for PET and CT images and nearest-neighbor interpolation for mask images. 

The randomized transforms were called at the start of every epoch. These included (i) randomly cropping cubic patches of dimensions $(N, N, N)$ from the images, where the cube was centered around a lesion voxel with probability $pos/(pos + neg)$, or around a background voxel with probability $neg/(pos+neg)$, (ii) translations in (-10, 10) voxels in 3D, (iii) axial rotations in $(-\pi/15, \pi/15)$, and (iv) random scaling by 1.1 in 3D. We set $neg=1$, and $pos$ and $N$ were chosen from the hyperparameter sets $\{1,2,4,6,8,10,12,14,16\}$ and $\{96, 128, 160, 192, 224, 256\}$ respectively for UNet \cite{training_patch_ablation}. After comprehensive ablation studies, $pos=2$ and $N = 224$ were found to be optimal for ResUNet. For other networks, $pos$ was set to 2, and the largest $N$ that could be accommodated into GPU memory during training was chosen (performance for different $N$ were not significantly different from each other, except $N=96$ which was significantly worse as compared to other values of $N$). Hence, SegResNet, DynUNet, and SwinUNETR were trained using $N=192$, 160, and 128, respectively. The augmented PET and CT patches were channel-concatenated to obtain the input to the network.


\subsubsection{Loss function, optimizer, and learning-rate scheduler}
\label{subsubsec:loss_function_optimizer_learning_rate_and_scheduler}
We employed the binary Dice loss $\mathcal{L}_{\text{Dice}}$ as the loss function given by,
\begin{equation}
    \mathcal{L}_{\text{Dice}} = 1 - \frac{1}{n_b}\sum_{i=1}^{n_b} \Bigg (\frac{2 \sum_{j=1}^{N^3} p_{ij} g_{ij} + \epsilon}{\sum_{j=1}^{N^3} (p_{ij} +  g_{ij}) + \eta} \Bigg)
\end{equation}
where, $p_{ij}$ and $g_{ij}$ are the $j^{th}$ voxel of the $i^{th}$ cropped patch of the predicted and GT segmentation masks, respectively in a mini-batch size $n_b = 1$ of the cropped patches and $N^3$ represents the total number of voxels in the patch of size $(N, N, N)$. Small constants $\epsilon = \eta = 10^{-5}$ were added to the numerator and denominator in $\mathcal{L}_{\text{Dice}}$ to ensure numerical stability during training. The loss was optimized using AdamW optimizer with a weight-decay of $10^{-5}$. Cosine annealing scheduler was used to decrease the learning rate from $2 \times 10^{-4}$ to $0$ in 500 epochs. The loss for an epoch was computed by averaging the $\mathcal{L}_{\text{Dice}}$ over all batches. The model with the highest mean DSC on the validation set was chosen for further evaluation.   

\subsubsection{Sliding window inference and postprocessing}
\label{subsubsec:sliding_window_inference_and_post_processing}
For inference, we employed only non-randomized transforms. The prediction was made on the 2-channel (PET and CT) whole-body images using the sliding-window technique with a window of dimension $(W, W, W)$, where $W$ was a hyperparameter chosen from $\{96, 128, 160, 192, 224, 256, 288\}$. The optimal $W$ was found to be 224, 192, 192, and 160 for ResUNet, SegResNet, DynUNet, and SwinUNETR, respectively. The test set predictions were resampled to the coordinates of the original GT masks for computing the evaluation metrics.

\subsection{Segmentation performance}
\label{subsec:segmentation_performance_app}

\subsubsection{Quantitative comparison between networks}
\label{subsubsec:quantitiative_comparison_between_networks}

\begin{table*}[h]
\centering
\resizebox{\linewidth}{!}{%
\LARGE
\begin{tabular}{clrrrrr}
\hline
 &
  \multicolumn{1}{c}{} &
  \multicolumn{5}{c}{\textbf{Cohorts (test set)}} \\ \cmidrule{3-7} 
 &
  \multicolumn{1}{c}{} &
  \multicolumn{4}{c}{\textbf{Internal}} &
  \multicolumn{1}{c}{\textbf{External}} \\ \cmidrule{3-7} 
\multirow{-3}{*}{\textbf{Networks}} &
  \multicolumn{1}{c}{\multirow{-3}{*}{\textbf{Metrics}}} &
  \multicolumn{1}{c}{\textbf{Overall}} &
  \multicolumn{1}{c}{\textbf{\begin{tabular}[c]{@{}c@{}}DLBCL-BCCV\\ (19 scans)\end{tabular}}} &
  \multicolumn{1}{c}{\textbf{\begin{tabular}[c]{@{}c@{}}PMBCL-BCCV\\ (25 scans)\end{tabular}}} &
  \multicolumn{1}{c}{\textbf{\begin{tabular}[c]{@{}c@{}}DLBCL-SMHS\\ (44 scans)\end{tabular}}} &
  \multicolumn{1}{c}{\textbf{\begin{tabular}[c]{@{}c@{}}AutoPET lymphoma\\ (145 scans)\end{tabular}}} \\ \hline
 &
  \cellcolor[HTML]{EFEFEF}DSC (mean) ($\uparrow$) &
  \cellcolor[HTML]{EFEFEF}0.60 $\pm$ 0.34 &
  \cellcolor[HTML]{EFEFEF}0.56 $\pm$ 0.35 &
  \cellcolor[HTML]{EFEFEF}0.49 $\pm$ 0.42 &
  \cellcolor[HTML]{EFEFEF}0.68 $\pm$ 0.27 &
  \cellcolor[HTML]{EFEFEF}0.56 $\pm$ 0.27 \\
 &
  \cellcolor[HTML]{FFFFFF}{\color[HTML]{212529} DSC ($\uparrow$)} &
  0.74 {[}0.26, 0.87{]} &
  \textbf{0.72 {[}0.24, 0.89{]}} &
  \textbf{0.74 {[}0.02, 0.9{]}} &
  0.78 {[}0.55, 0.87{]} &
  0.66 {[}0.33, 0.77{]} \\ \cline{2-7} 
 &
  \cellcolor[HTML]{EFEFEF}FPV ($\downarrow$) &
  \cellcolor[HTML]{EFEFEF}6.26 {[}0.68, 32.06{]} &
  \cellcolor[HTML]{EFEFEF}6.95 {[}0.0, 36.55{]} &
  \cellcolor[HTML]{EFEFEF}2.23 {[}0.13, 17.47{]} &
  \cellcolor[HTML]{EFEFEF}\textbf{8.71 {[}1.19, 34.1{]}} &
  \cellcolor[HTML]{EFEFEF}38.41 {[}15.86, 128.47{]} \\ 
\multirow{-4}{*}{ResUNet} &
  FNV ($\downarrow$) &
  0.0 {[}0.0, 5.05{]} &
  2.04 {[}0.0, 4.82{]} &
  \textbf{0.0 {[}0.0, 1.24{]}} &
  0.85 {[}0.0, 10.81{]} &
  \textbf{0.41 {[}0.0, 3.88{]}} \\ \hline
 &
  \cellcolor[HTML]{EFEFEF}DSC (mean) ($\uparrow$) &
  \cellcolor[HTML]{EFEFEF}0.60 $\pm$ 0.34 &
  \cellcolor[HTML]{EFEFEF}0.56 $\pm$ 0.37 &
  \cellcolor[HTML]{EFEFEF}0.51 $\pm$ 0.41 &
  \cellcolor[HTML]{EFEFEF}0.68 $\pm$ 0.27 &
  \cellcolor[HTML]{EFEFEF}0.58 $\pm$ 0.27 \\
 &
  DSC ($\uparrow$) &
  \textbf{0.76 {[}0.27, 0.88{]}} &
  0.71 {[}0.19, 0.9{]} &
  0.72 {[}0.01, 0.85{]} &
  \textbf{0.78 {[}0.62, 0.87{]}} &
  \textbf{0.68 {[}0.4, 0.78{]}} \\  \cline{2-7} 
 &
  \cellcolor[HTML]{EFEFEF}FPV ($\downarrow$) &
  \cellcolor[HTML]{EFEFEF}\textbf{4.55 {[}1.35, 31.51{]}} &
  \cellcolor[HTML]{EFEFEF}\textbf{5.78 {[}0.61, 19.97{]}} &
  \cellcolor[HTML]{EFEFEF}\textbf{2.15 {[}0.52, 7.18{]}} &
  \cellcolor[HTML]{EFEFEF}9.18 {[}2.08, 38.87{]} &
  \cellcolor[HTML]{EFEFEF}\textbf{21.46 {[}6.3, 66.44{]}} \\ 
\multirow{-4}{*}{SegResNet} &
  FNV ($\downarrow$) &
  0.0 {[}0.0, 7.55{]} &
  1.17 {[}0.0, 5.59{]} &
  0.0 {[}0.0, 6.3{]} &
  0.0 {[}0.0, 10.81{]} &
  0.5 {[}0.01, 3.88{]} \\ \hline
 &
  \cellcolor[HTML]{EFEFEF}DSC (mean) ($\uparrow$) &
  \cellcolor[HTML]{EFEFEF}0.56 $\pm$ 0.33 &
  \cellcolor[HTML]{EFEFEF}0.52 $\pm$ 0.37 &
  \cellcolor[HTML]{EFEFEF}0.45 $\pm$ 0.40 &
  \cellcolor[HTML]{EFEFEF}0.65 $\pm$ 0.24 &
  \cellcolor[HTML]{EFEFEF}0.55 $\pm$ 0.27 \\ 
 &
  DSC ($\uparrow$) &
  0.68 {[}0.3, 0.84{]} &
  0.69 {[}0.16, 0.87{]} &
  0.39 {[}0.02, 0.79{]} &
  0.73 {[}0.54, 0.83{]} &
  0.64 {[}0.33, 0.78{]} \\ \cline{2-7} 
 &
  \cellcolor[HTML]{EFEFEF}FPV ($\downarrow$) &
  \cellcolor[HTML]{EFEFEF}15.27 {[}3.27, 52.27{]} &
  \cellcolor[HTML]{EFEFEF}12.43 {[}1.24, 44.9{]} &
  \cellcolor[HTML]{EFEFEF}8.17 {[}2.39, 33.99{]} &
  \cellcolor[HTML]{EFEFEF}18.36 {[}8.85, 78.23{]} &
  \cellcolor[HTML]{EFEFEF}36.61 {[}13.61, 107.33{]} \\
\multirow{-4}{*}{DynUNet} &
  FNV ($\downarrow$) &
  0.0 {[}0.0, 7.31{]} &
  0.09 {[}0.0, 8.32{]} &
  \textbf{0.0 {[}0.0, 1.24{]}} &
  0.0 {[}0.0, 11.4{]} &
  0.42 {[}0.0, 4.17{]} \\ \hline
 &
  \cellcolor[HTML]{EFEFEF}DSC (mean) ($\uparrow$) &
  \cellcolor[HTML]{EFEFEF}0.57 $\pm$ 0.32 &
  \cellcolor[HTML]{EFEFEF}0.52 $\pm$ 0.35 &
  \cellcolor[HTML]{EFEFEF}0.46 $\pm$ 0.38 &
  \cellcolor[HTML]{EFEFEF}0.66 $\pm$ 0.24 &
  \cellcolor[HTML]{EFEFEF}0.53 $\pm$ 0.27 \\ 
 &
  DSC ($\uparrow$) &
  0.67 {[}0.4, 0.85{]} &
  0.48 {[}0.21, 0.89{]} &
  0.45 {[}0.01, 0.78{]} &
  0.75 {[}0.52, 0.85{]} &
  0.61 {[}0.33, 0.75{]} \\ \cline{2-7} 
 &
  \cellcolor[HTML]{EFEFEF}FPV ($\downarrow$) &
  \cellcolor[HTML]{EFEFEF}15.74 {[}3.86, 44.2{]} &
  \cellcolor[HTML]{EFEFEF}9.13 {[}3.19, 87.47{]} &
  \cellcolor[HTML]{EFEFEF}9.0 {[}3.87, 19.21{]} &
  \cellcolor[HTML]{EFEFEF}19.98 {[}7.85, 43.51{]} &
  \cellcolor[HTML]{EFEFEF}58.26 {[}19.35, 145.88{]} \\ 
\multirow{-4}{*}{SwinUNETR} &
  FNV ($\downarrow$) &
  \textbf{0.0 {[}0.0, 4.65{]}} &
  \textbf{0.09 {[}0.0, 3.39{]}} &
  \textbf{0.0 {[}0.0, 1.24{]}} &
  \textbf{0.0 {[}0.0, 8.83{]}} &
  0.58 {[}0.01, 3.52{]} \\ \hline
\end{tabular}}
\caption{Comparison of the four networks on the internal and external test sets evaluated via median values of patient-level DSC, FPV (in ml), and FNV (in ml). All the median values have been reported along with their IQRs. We also report the mean $\pm$ standard deviation for patient-level DSC.} 
\label{tab:final_segmentation_results}
\end{table*}

\begin{figure*}[]
\centering
\includegraphics[width=\textwidth]{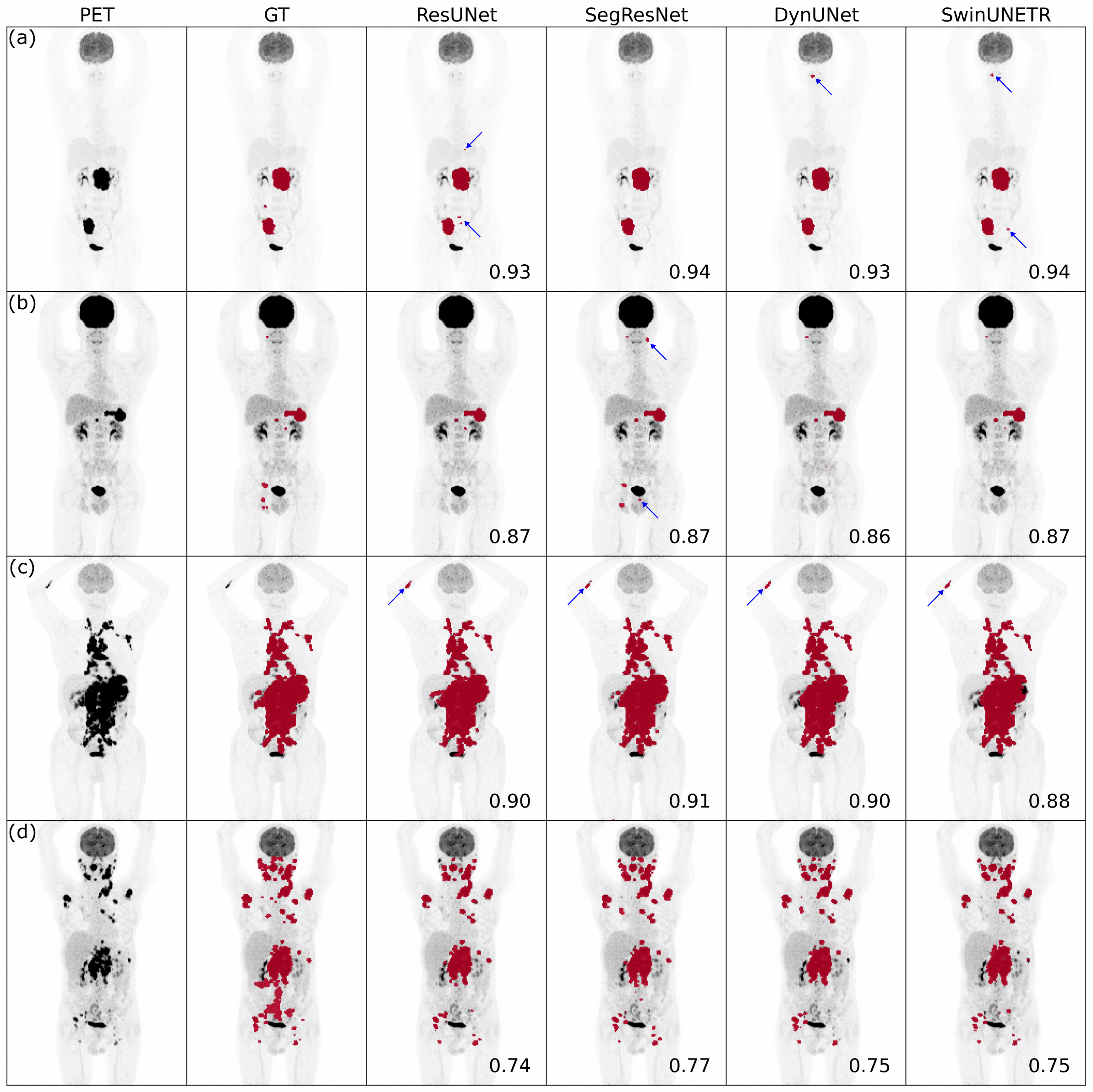}
\caption{Lesion segmentation by different networks shown in coronal maximum intensity projection views for four example cases where the \textbf{networks had similar performances}. The 3D DSC is shown at the bottom-right of each plot showing network predictions. Prominent FPVs have been indicated with blue arrows.}
\label{fig:images_similar_performance}
\end{figure*}

\begin{figure*}[]
\centering
\includegraphics[width=\textwidth]{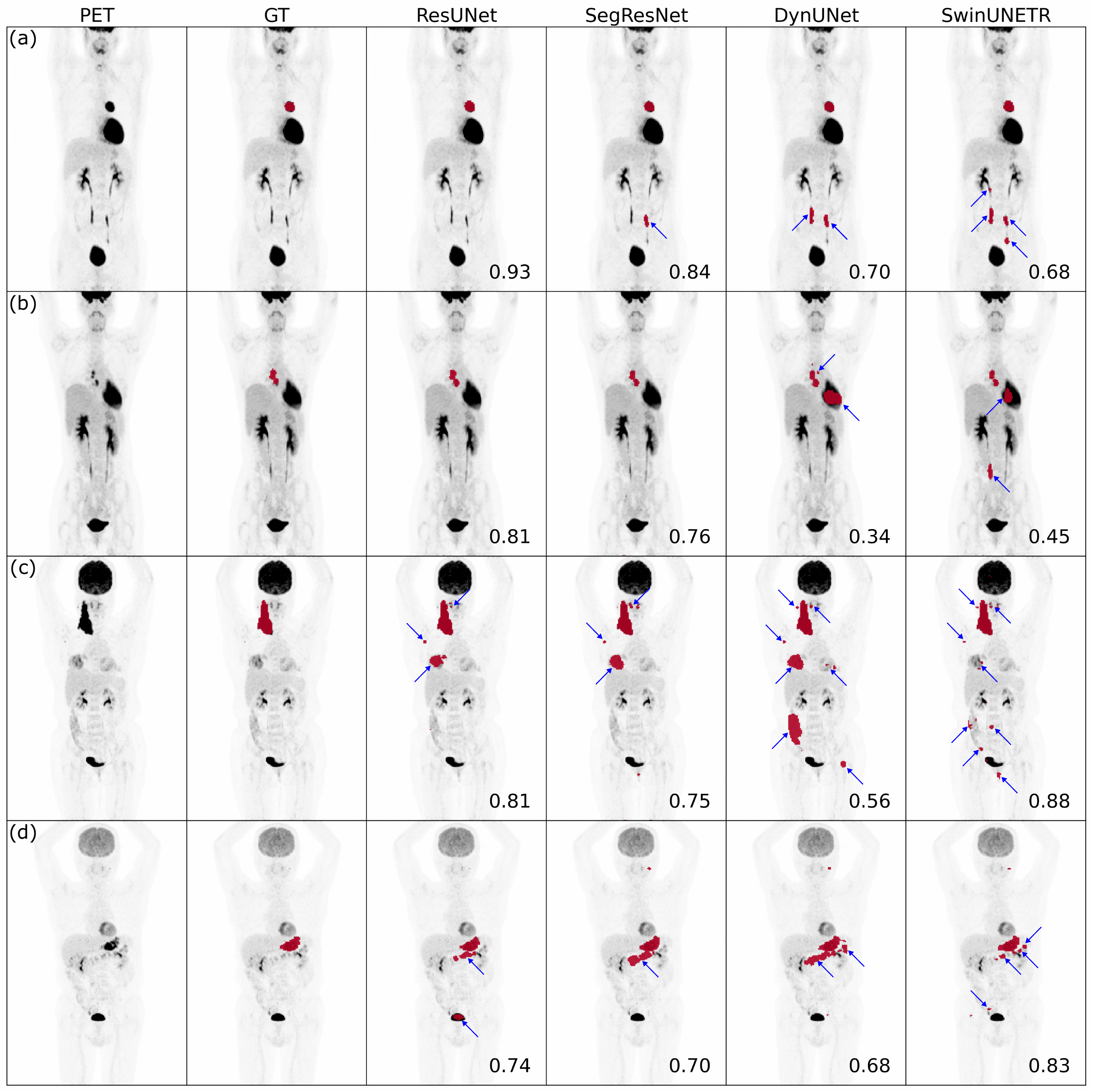}
\caption{Lesion segmentation by different networks shown in coronal maximum intensity projection views for four example cases where the \textbf{networks had dissimilar performances} often due to some of them predicting large false positives. The 3D DSC is shown at the bottom-right of each plot showing network predictions. Prominent FPVs have been indicated with blue arrows.}
\label{fig:images_dissimilar_performance}
\end{figure*}

Quantitative performance comparison between different networks are presented in \Cref{tab:final_segmentation_results}. Some visualizations showcasing networks predictions are provided in \Cref{fig:images_similar_performance} showing examples where the four networks had similar performances, and in \Cref{fig:images_dissimilar_performance} showing examples where the networks had dissimilar performances.\\

On the DSC metric, both SegResNet and UNet generalized well on the unseen external test set, with a drop in mean \& median DSC performance by 4\% \& 8\% and 2\% \& 8\%, respectively as compared to the internal test set. Although the median DSC of DynUNet and SwinUNETR are considerably lower than SegResNet and UNet on the internal test set (by about 6-9\%), these networks had even better generalizations with a drop in median DSC of only 4\% and 6\%, respectively, when going from internal to external testing. It is also worth noting that the DSC IQRs for all networks were larger on the internal test set as compared to the external test set. Also, all networks obtained a higher 75$^\text{th}$ quantile DSC on the internal test set as compared to the external test set, while obtaining a lower 25$^\text{th}$ quantile DSC on the internal test as compared to the external test set (except for SwinUNETR where this trend was reversed). Similarly, for different cohorts within the internal test set, all networks had the highest median and 25$^\text{th}$ quantile DSC on DLBCL-SMHS set. 


Secondly, SegResNet also had the best FPV on both internal and external test sets. Finally, SwinUNETR and UNet had the best median FNV on the internal and external test sets, respectively. On DLBCL-BCCV and DLBCL-SMHS, SwinUNETR had the best median FNV, while on PMBCL-BCCV, ResUNet, DynUNet, and SwinUNETR were tied, each with a median value of 0.0 [0.0, 1.24] ml. Interestingly, despite having a lower performance on DSC on both internal and external test sets (as compared to the best performing models), SwinUNETR had the best median FNV values across cohorts on the internal test set.

\subsubsection{Assessing the reproducibility of lesion measures via equivalence testing}
\label{subsubsec:assessing_the_reproducibility_of_lesion_measures_via_equivalence_testing}

\begin{figure*}[h!]
\subfigure[Internal test set]{
\includegraphics[width=0.5\linewidth]{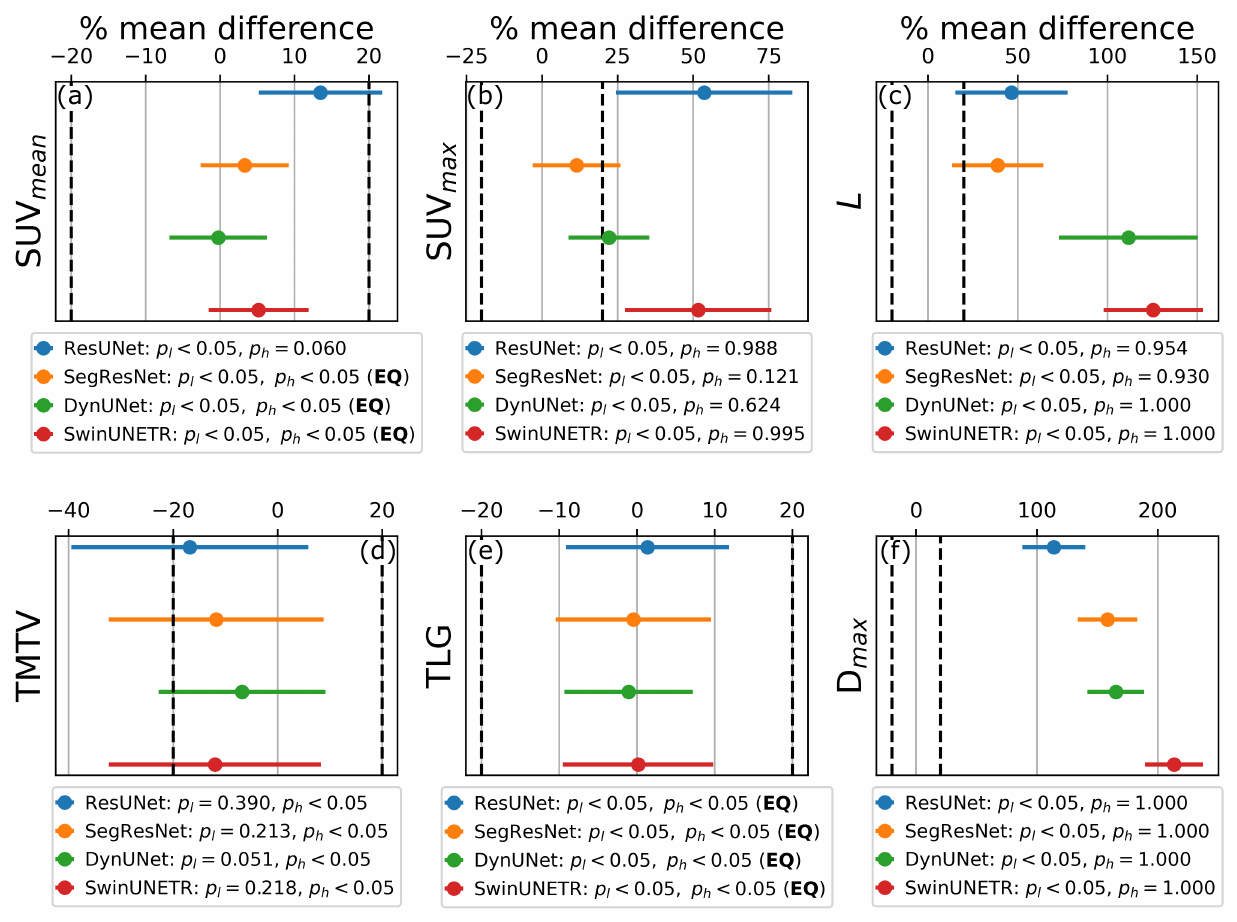}
}
\subfigure[External test set]{
\includegraphics[width=0.5\linewidth]{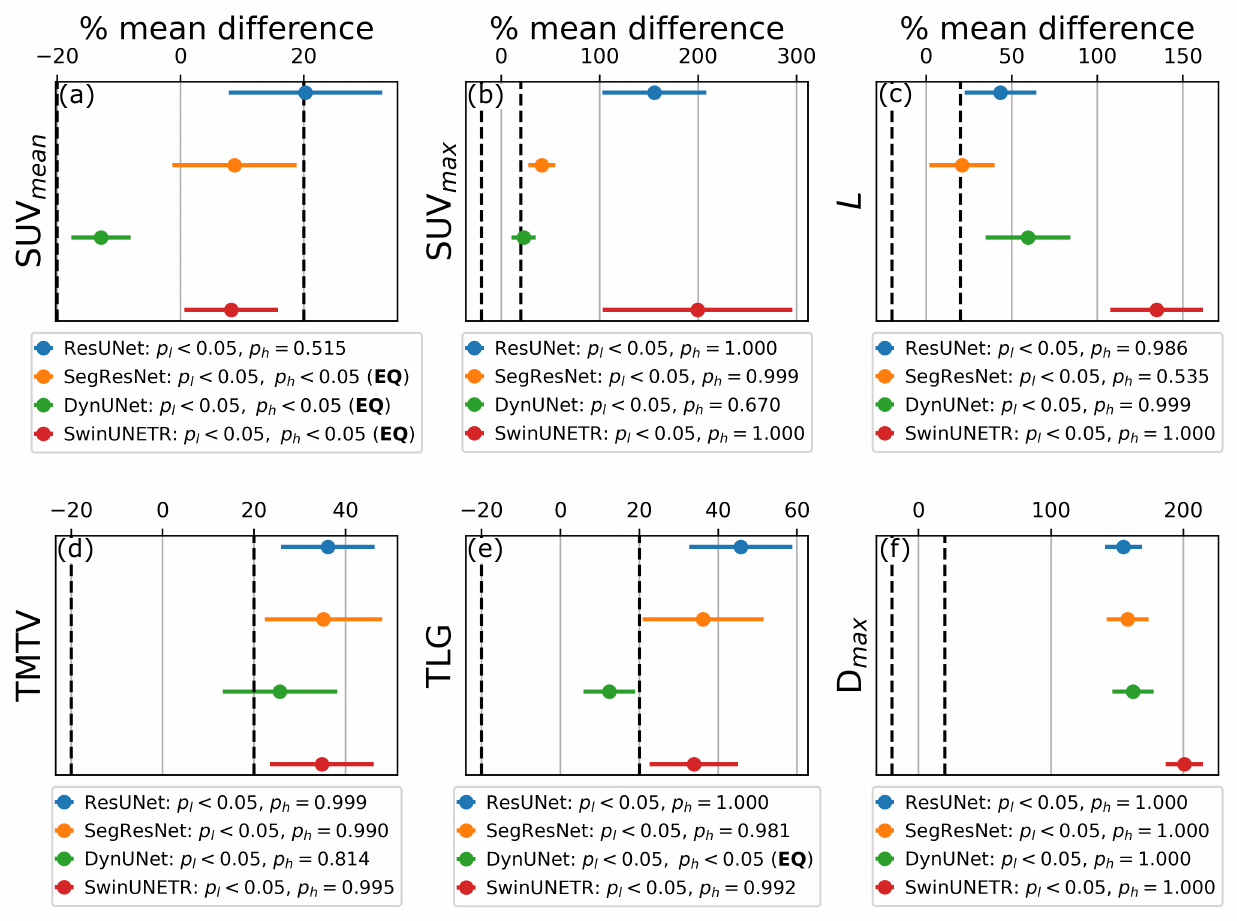}
}
\caption{Equivalence testing between the predicted and GT lesion measures via TOST procedure on the internal test set (left) with $N_\text{cases} = 88$ and the external test set (right) with  $N_\text{cases} = 145$ for the four networks. The forest plots show 95\% confidence interval for the \% mean difference between the predicted and GT measures. The region between the black dashed vertical lines at $\Delta = \pm 20\%$ shows the region of clinical equivalence. The labels \textbf{EQ} in the legends denote equivalence suggesting reproducibility with significance level $\alpha = 0.05$.}  
\label{fig:equivalence_testing}
\end{figure*}

For internal testing (\Cref{fig:equivalence_testing} (i)), SegResNet, DynUnet, and SwinUNETR showed reproducibility on SUV$_\text{mean}$, all networks showed reproducible on TLG, while no other measures were reproducible by any of the networks. For external testing as well (\Cref{fig:equivalence_testing} (ii)), only SegResNet, DynUnet, and SwinUNETR showed reproducibility on SUV$_\text{mean}$, only DynUNet on TLG, while all other measures were not reproducible by any of the networks. ResUNet, despite obtaining a high DSC (comparable to SegResNet) was not good at reproducing clinical metrics such as SUV$_\text{mean}$. This shows that the performance of networks in terms of DSC or other traditional segmentation metrics do not always reflect their adeptness at estimating the measures of clinical importance. Lesion measures such as $\text{SUV}_\text{max}$, $L$, TMTV and $\text{D}_\text{max}$ were hard to reproduce by the networks. $\text{SUV}_\text{max}$ was highly sensitive to incorrect false positive predictions in regions of physiological high SUV uptake. Similarly, $L$ was highly sensitive to incorrectly segmented disconnected components, and $\text{D}_\text{max}$ was highly sensitive to the presence of a false positive prediction far away from the GT segmentations.

\subsection{Networks segmentation performance distribution for different GT lesion measures on various test sets}
\label{subsubsec:lesion_measure_segregated_dsc_for_other_networks} 
\begin{figure*}[h]
\centering
\includegraphics[width=0.95\linewidth]{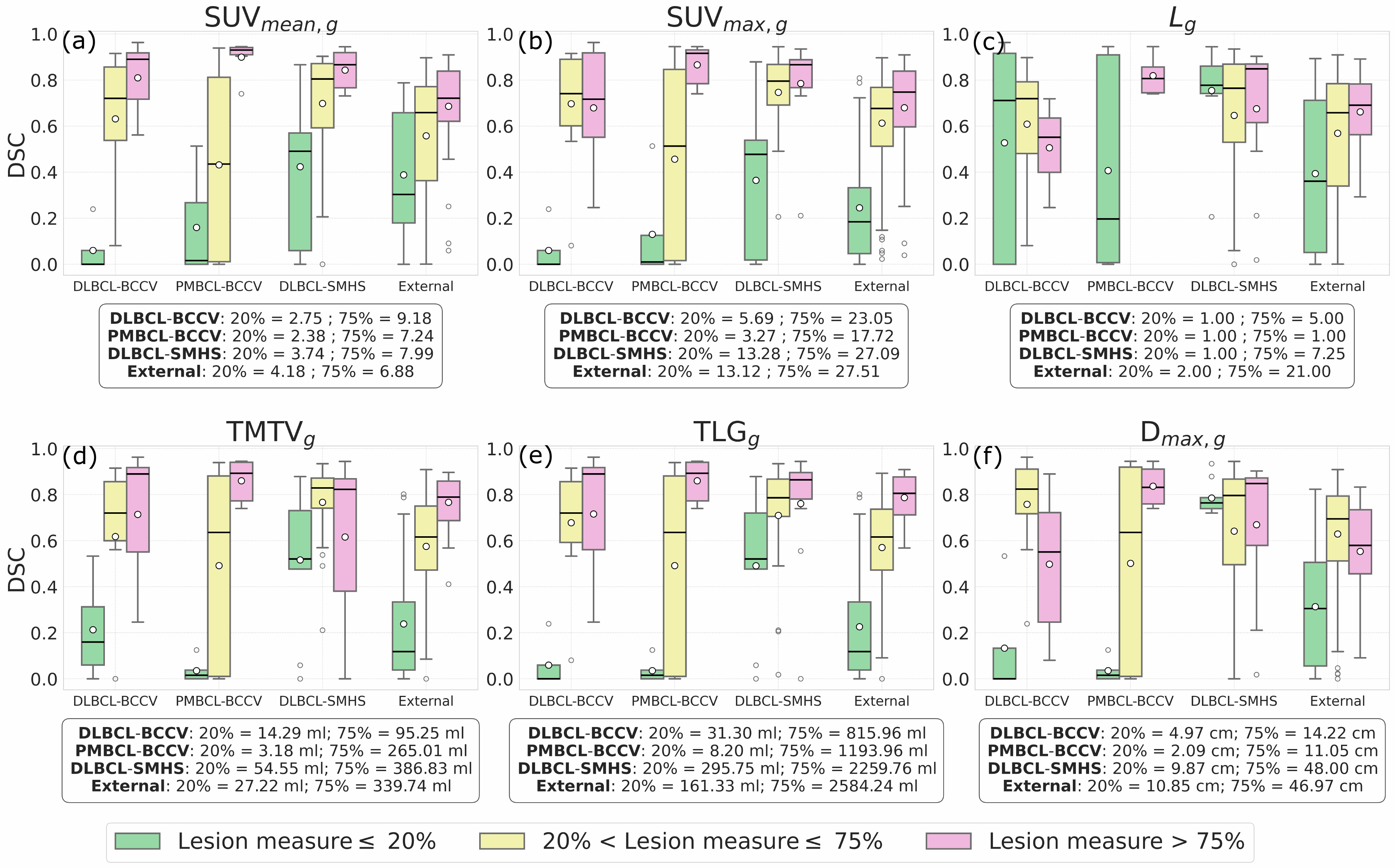}
\caption{\textbf{ResUNet} performance (DSC) distribution for different GT lesion measures on various test sets. For each test set, the DSC measure distributions have been presented as boxplots in three different categories, (i) Lesion measure $\leq$ 20\%tile, (ii)  20\%tile $<$ Lesion measure $\leq$ 75\%tile, (iii) Lesion measure $>$ 75\%tile. The mean and median values for each box have been represented as white circles and black horizontal lines, respectively. The boxes below each plot show the value of the 20\%tile and 75\%tile lesion measure on each of the test sets.}
\label{fig:lesion_measures_segregated_dsc_metrics_unet}
\end{figure*}

\begin{figure*}[h]
\centering
\includegraphics[width=0.95\linewidth]{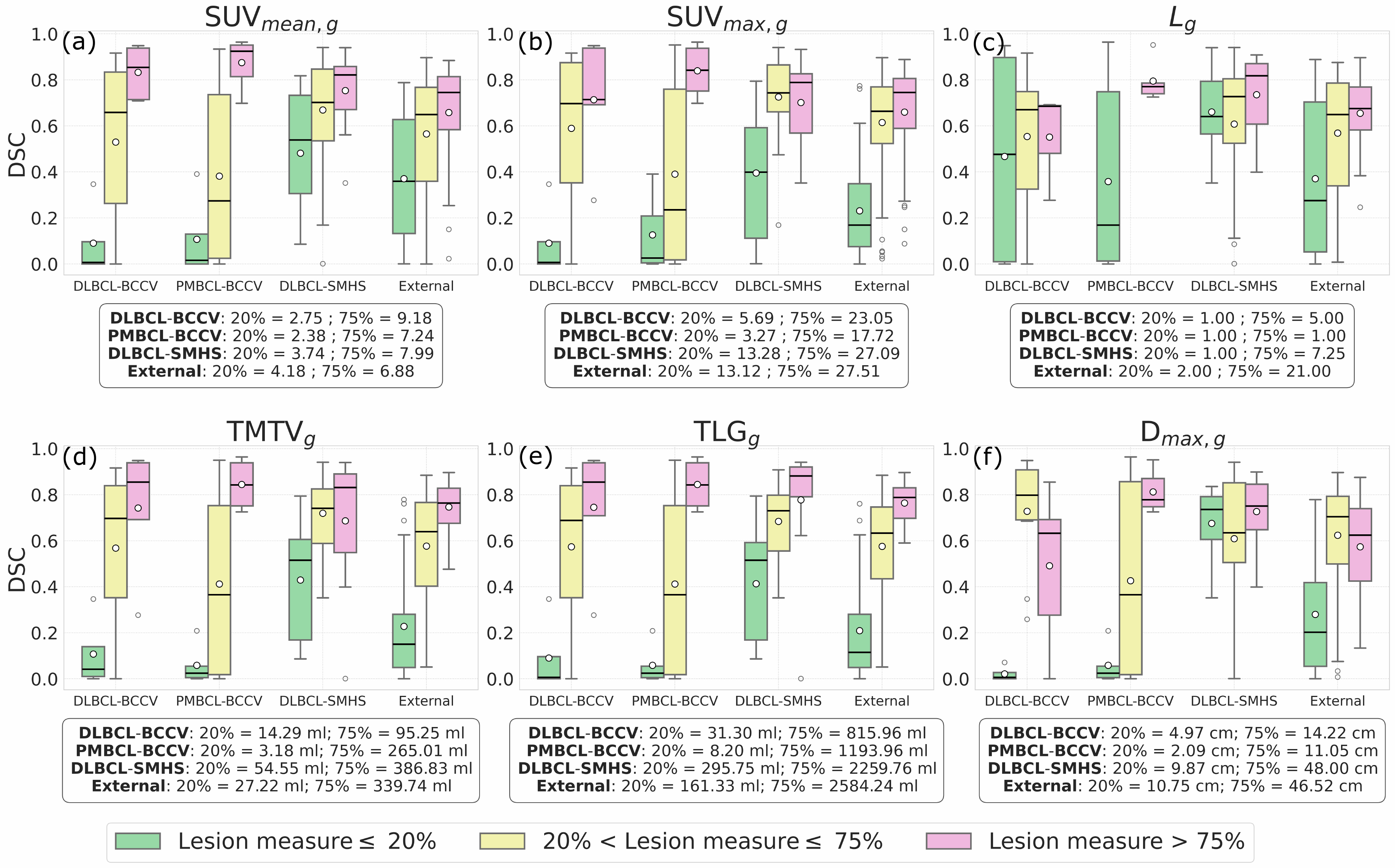}
\caption{\textbf{DynUNet} performance (DSC) distribution for different GT lesion measures on various test sets. For each test set, the DSC measure distributions have been presented as boxplots in three different categories, (i) Lesion measure $\leq$ 20\%tile, (ii)  20\%tile $<$ Lesion measure $\leq$ 75\%tile, (iii) Lesion measure $>$ 75\%tile. The mean and median values for each box have been represented as white circles and black horizontal lines, respectively. The boxes below each plot show the value of the 20\%tile and 75\%tile lesion measure on each of the test sets.}
\label{fig:lesion_measures_segregated_dsc_metrics_dynunet}
\end{figure*}

\begin{figure*}[h]
\centering
\includegraphics[width=0.95\linewidth]{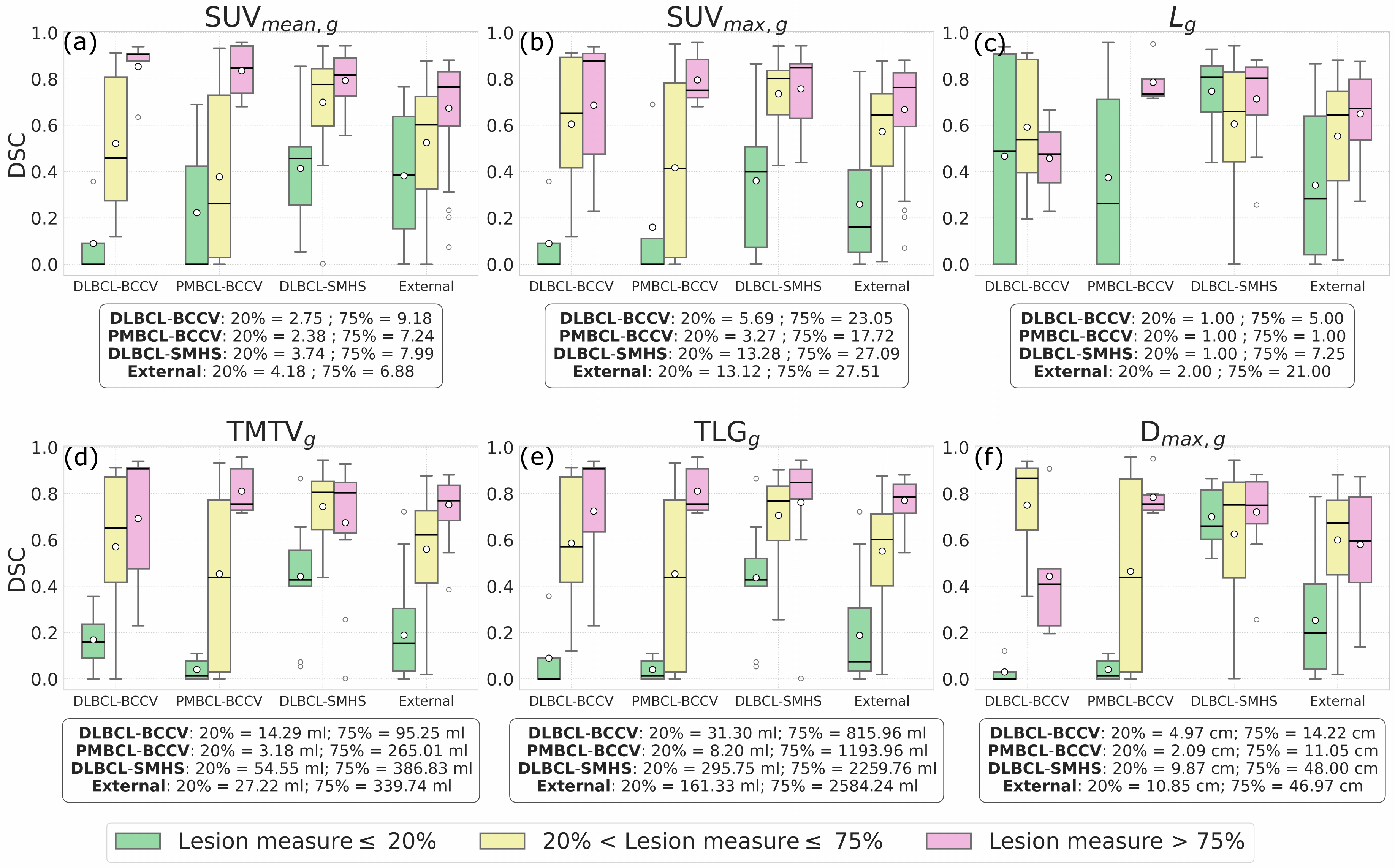}
\caption{\textbf{SwinUNETR} performance (DSC) distribution for different GT lesion measures on various test sets. For each test set, the DSC measure distributions have been presented as boxplots in three different categories, (i) Lesion measure $\leq$ 20\%tile, (ii) 20\%tile $<$ Lesion measure $\leq$ 75\%tile, (iii) Lesion measure $>$ 75\%tile. The mean and median values for each box have been represented as white circles and black horizontal lines, respectively. The boxes below each plot show the value of the 20\%tile and 75\%tile lesion measure on each of the test sets.}
\label{fig:lesion_measures_segregated_dsc_metrics_swinunetr}
\end{figure*}

\Cref{fig:lesion_measures_segregated_dsc_metrics_unet,fig:lesion_measures_segregated_dsc_metrics_dynunet,fig:lesion_measures_segregated_dsc_metrics_swinunetr} show the distribution of DSC for ResUNet, DynUNet and SwinUNETR, respectively, segregated by the test cohorts categorized within specific ranges of the lesion measure values ($\leq$20\%, 20\%-75\%, and >75\%) for all lesion measures. The various performance trends as a function of different ranges of lesion measures in these figures resemble those of SegResNet shown in \Cref{fig:lesion_measures_segregated_dsc_metrics_segresnet} in \Cref{subsubsec:dsc_vs_biomarkers_combined}.

When the trained models were applied to the external test set, we observed noticeably larger errors in derived lesion measures, TLG, TMTV, and $L$, even though the volumetric Dice coefficient remained broadly stable. \Cref{tab:lesion_measures_internal_external} reveals a pronounced shift in disease burden between cohorts: the external patients contained a far greater number of lesions (median $L$=9 vs 2) and larger dissemination (median $D_\text{max}$=24.6 cm vs 12.5 cm), together with higher SUV$_\text{max}$, TMTV and TLG. These differences magnify two error modes intrinsic to our pipeline: (i) each false-positive blob disproportionately inflates $L$ and $D_\text{max}$ in patients who already present dozens of lesions; and (ii) small boundary deviations around bulky or highly avid lesions translate into sizable absolute errors in TMTV and TLG. Future work could therefore focus on domain-adaptation strategies, lesion-aware loss functions that penalize missed or spurious connected components, and cascaded “detect-then-refine” architectures that first localize candidate foci before boundary delineation. In addition, aggregating labels from multiple readers or incorporating human-in-the-loop review for uncertain detections may further stabilize patient-level metrics across institutions and scanner protocols.

\begin{table*}[]
\centering
\resizebox{0.80\textwidth}{!}{%
\begin{tabular}{@{}ccccc@{}}
\toprule
\multirow{2}{*}{GT lesion measure} & \multicolumn{2}{c}{Internal test set}       & \multicolumn{2}{c}{External test set}       \\ \cmidrule(l){2-5} 
          & mean $\pm$ std  & median [range{]}   & mean $\pm$ std  & median [range{]}   \\ \midrule
SUV$_\text{mean}$   & 5.7 $\pm$ 3.4   & 4.9 [0.8-17.1{]}   & 6.1 $\pm$ 2.6   & 5.4 {[}2.2-16.7{]}   \\
SUV$_\text{max}$    & 17.9 $\pm$ 15.2 & 16.8 {[}1.5-123.8{]} & 22.8 $\pm$ 12.5 & 20.2 {[}2.7-101.0{]} \\
$L$        & 5.9 $\pm$ 10.6  & 2 {[}1-65{]}         & 22.8 $\pm$ 43.6 & 9 {[}1-325{]}        \\
TMTV (ml)                          & 280.8 $\pm$ 488.2   & 79.9 {[}0.5-2299.8{]}   & 297.4 $\pm$ 393.1   & 162.6 {[}0.1-2481.2{]}  \\
TLG (ml)                           & 1672.5 $\pm$ 2854.3 & 581.9 {[}1.3-14810.1{]} & 1971.1 $\pm$ 2834.2 & 952.3 {[}0.2-17067.6{]} \\
$D_\text{max}$ (cm) & 21.9 $\pm$ 23.3 & 12.5 [0.9-101.1] & 30.7 $\pm$ 21.4 & 24.6 {[}1.3-92.5{]}  \\ \bottomrule
\end{tabular}%
}
\caption{Table showing the difference between the ground truth distribution of various lesion measures such as SUV$_\text{mean}$, SUV$_\text{max}$, $L$, TMTV, TLG, and $D_\text{max}$. This shows that the external test set had a considerably large number of lesions (higher $L$), a higher TLG, as well as higher $D_\text{max}$ as compared to the internal test dataset, which are some of the contributing factors to suboptimal model performance in the reproducibility of lesion measures going from internal to external testing. Additionally, lesion measures like $L$ and $D_\text{max}$ are noticeably sensitive to the presence of false positives which makes it harder for the network to reproduce these metrics faithfully.}
\label{tab:lesion_measures_internal_external}
\end{table*}

\end{document}